\documentclass{article}
\usepackage{iclr2027_conference,times}

\usepackage{hyperref}
\usepackage{url}

\usepackage[utf8]{inputenc}
\usepackage[T1]{fontenc}
\usepackage{booktabs}
\usepackage{microtype}
\usepackage[table]{xcolor}
\usepackage[most]{tcolorbox}
\tcbset{breakable} %
\usepackage{enumitem}

\usepackage{graphicx}
\usepackage{amsmath}
\usepackage{amssymb}
\usepackage{subcaption}
\usepackage{multirow}

\newcommand{\eg}{\emph{e.g.}}
\newcommand{\nickname}{\textsc{ThinkDrag}}

\title{ThinkDrag: Semantic Drag-Based\\Image Editing with Visual Reasoning}

\makeatletter\@ifundefined{iflatexml}{\newif\iflatexml}{}\makeatother
\iflatexml
  \usepackage{authblk}
  \author{Armando Fortes}\author{Tianyi Wei}\author{Chindanai Trakantannarong}\author{Shangchen Zhou}\author{Xingang Pan}
  \affil{S-Lab, Nanyang Technological University, Singapore}
\else
\author{%
  Armando Fortes, Tianyi Wei, Chindanai Trakantannarong, Shangchen Zhou, Xingang Pan\\
  S-Lab, Nanyang Technological University, Singapore\\
  \url{https://atfortes.github.io/projects/thinkdrag/}
}
\fi

\iclrfinalcopy %
\begin{document}

\setlength{\tabcolsep}{0pt} %
\maketitle
\setlength{\tabcolsep}{6pt}
\lhead{Preprint}
\iflatexml
\begin{center}Project page: \url{https://atfortes.github.io/projects/thinkdrag/}\end{center}
\fi

\vspace{-10pt}
\begin{abstract}

Drag-based image editing provides an intuitive interface for spatially controlled image manipulation by allowing users to specify handle and target points. Existing methods have made substantial progress through optimization-based, one-step, and feed-forward formulations, but they often interpret drag instructions as geometric displacements. This limits their effectiveness when the desired edit depends on interpreting image content and inferring a plausible semantic transformation from sparse point constraints. We introduce \nickname, a unified multimodal framework for drag-based image editing with visual reasoning. \nickname\,is trained to associate point constraints with meaningful object- and scene-level transformations, and can optionally produce an explicit reasoning trace that interprets the intended edit before image generation. This reasoning trace improves interpretability, particularly for ambiguous drags. Drag Endpoint Markers, motivated by an analysis of cross-image attention, provide an explicit correspondence signal that grounds each drag in image space. To support this framework, we construct a supervised dataset of semantic drag transformations paired with reasoning traces and introduce DragBench++, a benchmark targeting challenging drag-based editing scenarios with reference edit solutions. Experiments show that \nickname\,achieves state-of-the-art performance, improving generation quality and plausibility while maintaining competitive point-following precision.

\end{abstract}

\vspace{-10pt}
\section{Introduction}
\label{sec:intro}

Image editing is a fundamental problem in generative modeling, where a model must modify visual content while keeping the rest of the scene intact~\citep{meng2022sdedit, hertzPrompttoPromptImageEditing2023, brooksInstructPix2PixLearningFollow2023}. Beyond content creation, it also tests how well a model connects visual understanding with controllable synthesis~\citep{dengEmergingPropertiesUnified2025}. Text-guided editing has made image manipulation more accessible, but language alone often provides limited spatial precision. Many edits are easier to specify by directly indicating where visual content should move, such as repositioning an object part or changing a pose. Drag-based image editing~\citep{panDragYourGan2023, shi2024dragdiffusion} provides this interface by allowing users to specify pairs of handle and target points, after which the model edits the image by moving the content associated with each handle point toward its target.

Early methods such as DragGAN~\citep{panDragYourGan2023} demonstrated drag-based editing with GAN priors, while diffusion-based methods extended it to broader visual domains~\citep{shi2024dragdiffusion, moudragondiffusion}. Existing approaches typically rely on latent optimization~\citep{zhanggooddrag, xiadraglora}, one-step latent editing~\citep{zhao2024fastdrag, pu2026dragging}, or feed-forward generation conditioned on point coordinates~\citep{shin2024instantdrag, shi2025lightningdrag}. However, they largely treat drag instructions as spatial displacements and rely on the base generative prior for plausibility. This is limiting because the same handle--target displacement can correspond to different valid edits depending on the object and on what the user intends. For example, moving a point on an articulated object may require rotating a part rather than translating a local patch.

Recent unified multimodal models~\citep{zhoutransfusion, xieshow, dengEmergingPropertiesUnified2025} provide a new opportunity for drag-based image editing. Unlike conventional image editors designed primarily around synthesis~\citep{brooksInstructPix2PixLearningFollow2023, wu2025qwen, labs2025flux}, these models couple visual and language understanding with image generation in a single framework. This extends the intuition behind DragGAN~\citep{panDragYourGan2023}: if a generative prior enables drag editing within a limited domain, then unified multimodal models can provide a broader prior for arbitrary images by drawing on their knowledge of the visual world. Such a prior is particularly suitable for drag-based editing, where a point instruction is not simply a displacement vector but an under-specified editing request. Executing it correctly requires understanding what the scene contains and how its objects can move or change.
We therefore formulate drag-based editing as semantic interpretation coupled with correspondence-grounded generation.

\begin{figure}[t]
\includegraphics[width=\linewidth]{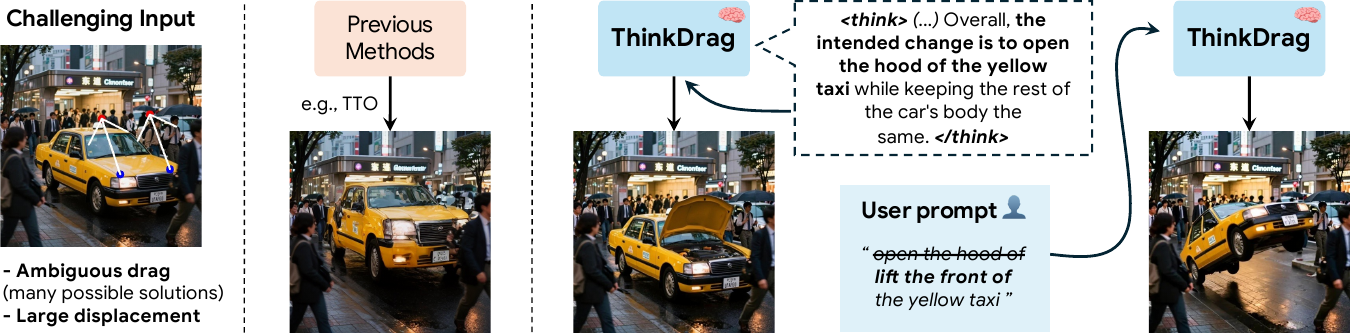}
    \caption{\textsc{ThinkDrag} handles challenging and ambiguous drags, optionally making its interpretation explicit as a reasoning trace that the user can refine, while the drags keep the edit spatially grounded.}
    \label{fig:teaser}
    \vspace{-3pt}
\end{figure}

In this work, we introduce \nickname, a unified multimodal framework for drag-based image editing with visual reasoning. Given an input image and drag instructions, \nickname\,generates an edited image that satisfies the point constraints while preserving the semantic structure of the scene. This is enabled by our new dataset of high-quality semantic drag edits, which teach the model to associate sparse point constraints with plausible object- and scene-level transformations, paired with reasoning traces that teach it to interpret the intended edit before generation. To execute drags precisely, we further introduce Drag Endpoint Markers (DEM). Motivated by the observation that each region of the edited image attends predominantly to the source region from which its content originates, DEM explicitly encodes the correspondence between each handle and its target. As illustrated in Figure~\ref{fig:teaser}, \nickname\,uses visual reasoning to interpret ambiguous drag instructions and can further incorporate user text when additional semantic guidance is provided. The model can edit directly, or first reason when the edit is ambiguous or requires higher-level scene understanding.

We evaluate \nickname\,on standard drag-editing benchmarks and introduce DragBench++, a new benchmark targeting more challenging scenarios that require higher-level scene understanding and semantic interpretation of point constraints. Extensive experiments show that \nickname\,achieves state-of-the-art performance, improving generation quality while maintaining competitive point-following precision. The direct editing mode already performs strongly, reflecting the value of semantic drag supervision and unified multimodal modeling. The reasoning mode complements direct editing by making the model's interpretation explicit, and it is particularly useful for ambiguous multi-drag edits. In a user study, participants prefer \nickname\,to prior methods in 80\% of the overall-preference votes on DragBench++ and 61\% on DragBench.

Our contributions are summarized as follows:
\begin{itemize}[leftmargin=25pt]
    \item We introduce \nickname, a unified multimodal framework for drag-based image editing with visual reasoning, interpreting drag instructions with scene semantics and object structure.
    \item We construct, to our knowledge, the first supervised dataset of high-quality semantic drag transformations with reasoning traces for training drag-based image editing models.
    \item We propose DragBench++, a benchmark targeting more challenging and semantically ambiguous drag-based editing scenarios, with reference edit solutions.
    \item We achieve state-of-the-art results on drag-based editing benchmarks and support joint text-drag conditioning, where text disambiguates user intent and drags provide precise spatial control.
\end{itemize}

\section{Related Work}
\label{sec:related}

\subsection{Unified Multimodal Understanding and Generation}

Multimodal understanding and image generation have traditionally followed different modeling paradigms, with autoregressive vision-language models driving perception and reasoning~\citep{liuVisualInstructionTuning2023, liu2024improved, bai2023qwen}, and diffusion models dominating high-fidelity synthesis~\citep{hoDenoisingDiffusionProbabilistic2020, songDenoisingDiffusionImplicit2021, dhariwalDiffusionModelsBeat2021}. Recent unified multimodal models combine both capabilities in a single framework, either by tokenizing images and modeling them jointly with text through next-token prediction~\citep{team2024chameleon, wang2024emu3, wu2025janus, chen2025janus} or by coupling autoregressive language modeling with diffusion-based visual generation~\citep{xie2025show, zhoutransfusion, dengEmergingPropertiesUnified2025, liao2025mogao}. Multimodal chain-of-thought~\citep{wang2025multimodal, shao2024visual} and generator-coupled reasoning~\citep{qin2026unicot, he2026re} further show a growing ability to connect visual perception, language reasoning, and image synthesis, but their use for spatially grounded image manipulation remains underexplored.

\subsection{Drag-based Image Editing}

Drag-based image editing provides intuitive image manipulation by dragging user-specified handle points to target positions. DragGAN~\citep{panDragYourGan2023} introduced this paradigm through latent optimization with motion supervision and point tracking, but its GAN prior limits generalization to arbitrary real images. Diffusion-based methods extend drag editing to broader domains, mostly through test-time optimization~\citep{shi2024dragdiffusion, moudragondiffusion, hou2024easydrag, ling2024freedrag, liu2024drag, cui2024localize, zhanggooddrag, jiangclipdrag, choi2025dragtext, xiadraglora, zhou2026dragnext, lin2025gdragtowards} or faster one-step and training-free alternatives~\citep{nieblessing, zhao2024fastdrag, lu2024regiondrag, lu2025inpaint4drag, yin2026lazydrag, pu2026dragging, Jhong_2026_WACV, he2025contextdrag}. Most related to our work are feed-forward methods that frame drag-based editing as conditional generation. InstantDrag~\citep{shin2024instantdrag} and LightningDrag~\citep{shi2025lightningdrag} synthesize edited images directly from point or motion conditions derived from videos, treating drags as geometric constraints, while \citet{cui2024localize} infer the intended edit with an LLM at test time. In contrast, \nickname\,interprets drag instructions within a unified multimodal model and grounds edits through an explicit correspondence signal in image space, learned from semantic drag edits.

\section{\nickname}
\label{sec:method}

\subsection{Dataset Construction}
\label{sec:dataset}

\begin{figure}[t]
\includegraphics[width=\linewidth]{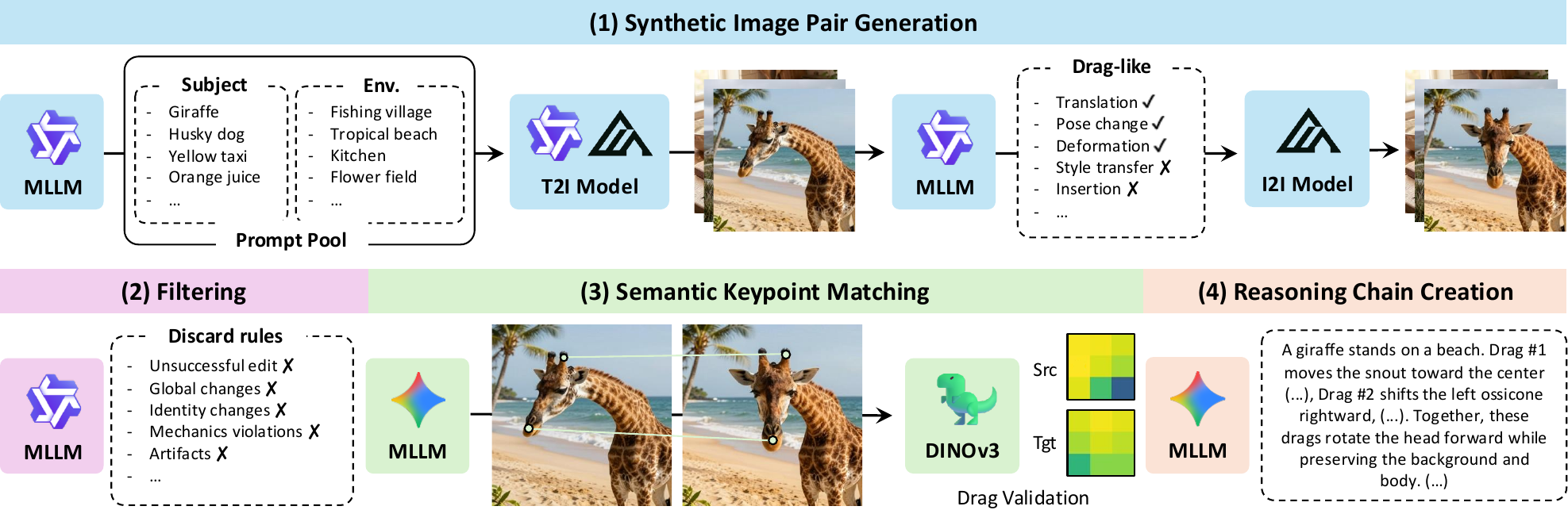}
    \caption{Dataset construction pipeline of the \textsc{ThinkDrag}\,dataset.}
    \label{fig:data_pipe}
\end{figure}

Prior work~\citep{shin2024instantdrag, shi2025lightningdrag} has shown that videos provide a natural source of paired supervision for drag-based image editing. These pipelines typically sample source and target images from a temporal window and recover motion using optical flow or point tracking.
However, video-derived supervision has several limitations. Extracted correspondences are primarily geometric and may not reflect semantically meaningful edit intent, optical flow can be unreliable under large displacement or occlusion, and natural videos often contain camera motion or background variation that corrupts the desired edit supervision. We therefore construct the \nickname\,dataset, a new drag-based image editing dataset that emphasizes high-quality image pairs and semantically meaningful point instructions. Figure~\ref{fig:data_pipe} gives an overview of the pipeline, described below.

\paragraph{Synthetic Image Pair Generation.}
Drag-based image editing is best suited to transformations that can be expressed through spatial motion of existing visual content, such as object translation, pose change, non-rigid deformation, and local geometric warping. It is less appropriate for edits involving semantic discontinuities (\eg, object insertion and removal, or global style transfer). Based on this distinction, we construct synthetic image pairs through an automated MLLM-assisted pipeline. We first use an MLLM~\citep{bai2025qwen3} to generate diverse source-image prompts from a pool of subjects and environments, and synthesize source images using state-of-the-art text-to-image models, including Qwen-Image~\citep{wu2025qwen}, FLUX.2~\citep{flux2klein9b}, and Z-Image~\citep{cai2025z}. Given each source image, the MLLM proposes image-specific edit prompts constrained to be compatible with drag-like transformations.
The source image and edit prompt are then passed to the editing model FLUX.2~\citep{flux2klein9b, flux2devturbo} to synthesize the target image, and an MLLM-based binary filter discards low-quality or invalid pairs.

\begin{figure}[t]
\includegraphics[width=\linewidth]{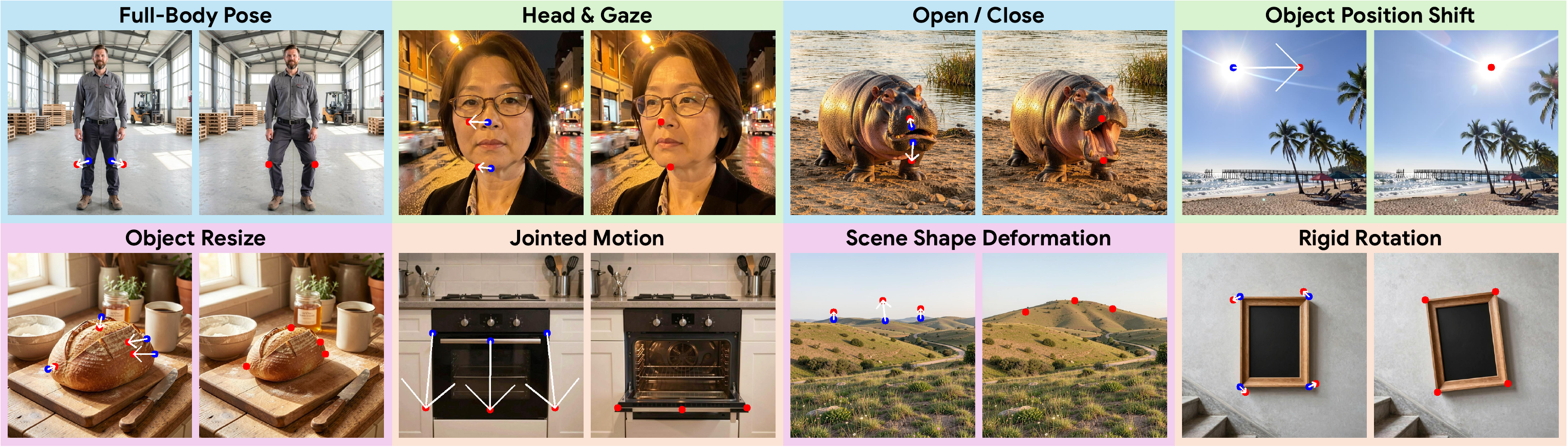}
    \caption{Diverse examples of the \textsc{ThinkDrag}\,dataset.}
    \vspace{-10pt}
    \label{fig:data}
\end{figure}

\paragraph{Semantic Keypoint Matching.}
From the filtered pairs, we recover each edit as semantic drag instructions. We prompt Gemini 3 Pro~\citep{gemini3} to identify keypoints that explain the transformation between the source and target images. Each drag consists of a handle point, a target point, and a semantic description of the matched element in both images. This converts the synthetic pair into drag-based supervision while preserving the semantic role of each correspondence. Unlike video-derived annotations from optical flow, this procedure targets the user-facing edit intent rather than only low-level motion. Since MLLMs can hallucinate incorrect matches, we validate the predicted correspondences using DINOv3~\citep{simeoniDINOv32025} features and manual inspection.

\paragraph{Reasoning Chain Creation.}
Finally, we generate reasoning chains for each drag-edit example using Gemini 3 Flash~\citep{gemini3}. The dataset construction pipeline makes the intended edit recoverable by construction: the MLLM has access to the source image, edited image, edit prompt, and semantic keypoint correspondence annotations. We use this information to prompt the MLLM to produce a reasoning chain that explains the source scene, interprets each drag in semantic terms, and summarizes the intended transformation while specifying which regions should remain unchanged. Since each source--target pair can be used bidirectionally, we generate reasoning chains for both edit directions. These traces provide supervision for reasoning-driven drag-based editing.

\paragraph{Dataset Statistics.}
Starting from approximately 193K generated pairs, MLLM filtering retains 97K, keypoint matching 50K, and verification yields the final dataset of approximately 11K source--target image pairs from 7K source images. Since each pair can be used in both directions, by swapping the source and target images and reversing the drag endpoints, this yields approximately 22K drag-based editing samples.
During training, we further apply standard spatial augmentations such as horizontal flips and random crops. Figure~\ref{fig:data} shows how the dataset spans diverse semantic drag transformations. Samples contain 2.15 drags on average (74\% with multiple drags) and reasoning traces of 123 words on average. Edit category distribution and per-stage details are provided in Appendix~\ref{app:dataset}.

\subsection{Drag Conditioning in a Unified Multimodal Model}

\begin{figure}[t]
\includegraphics[width=\linewidth]{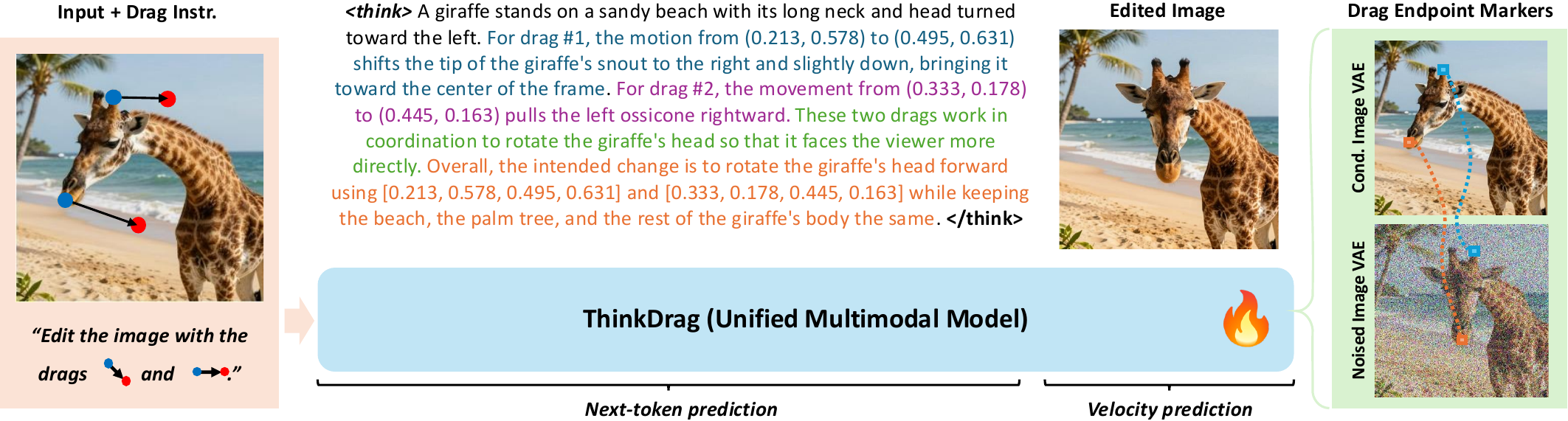}
    \caption{Overview of the \nickname\,unified multimodal model pipeline.}
    \vspace{-10pt}
    \label{fig:method}
\end{figure}

We build on BAGEL~\citep{dengEmergingPropertiesUnified2025}, a unified multimodal model with a mixture-of-transformers architecture~\citep{liang2025mixture}, and adapt it to drag-based editing by representing drag instructions through its existing multimodal interface. As shown in Figure~\ref{fig:method}, rather than introducing a separate point encoder or treating drags only as low-level geometric inputs, we express each drag through both structured text tokens and spatial endpoint markers. The text tokens expose the paired source and target coordinates, normalized to $[0,1]$ with three decimals, to the model in a form compatible with its language and visual understanding capabilities, allowing the drag instruction to be processed together with the image context and used for reasoning through the model's \emph{understanding expert}. The endpoint markers complement this representation by injecting the same drag information at the corresponding VAE token locations, providing explicit spatial anchors for image generation through the model's \emph{generation expert}. This design is motivated by the observation that each region of the generated image attends predominantly to the source region from which its content originates (Figure~\ref{fig:dem_ablation}). This suggests that a drag is more likely to be executed correctly when the target location attends to its handle, a correspondence that textual coordinates convey only indirectly.

\paragraph{Drag Endpoint Markers.}
Let the drag instruction be $\mathcal{D}=\{(h_k,t_k)\}_{k=1}^{K}$, where each drag $k$ consists of a handle point $h_k=(x^h_k,y^h_k)$ and a target point $t_k=(x^t_k,y^t_k)$. We associate each drag index $k$ with a marker vector $\mathbf{m}_k=\mathbf{m}_{\mathrm{shared}}+\mathbf{e}_k$, where $\mathbf{m}_{\mathrm{shared}}\in\mathbb{R}^{d}$ is shared across all drags, initialized to zero and learned, and $\mathbf{e}_k\in\mathbb{R}^{d}$ is a fixed orthogonal slot embedding that preserves the identity of drag $k$. Let $\mathbf{z}^{(\ell)}_{\mathrm{src}}$ and $\mathbf{z}^{(\ell)}_{\mathrm{tgt}}$ denote the source and target VAE hidden features before layer $\ell$. We inject drag markers by applying additive updates at the corresponding spatial locations:
\begin{align}
    \mathbf{z}^{(\ell)}_{\mathrm{src}}(h_k)
    &\leftarrow
    \mathbf{z}^{(\ell)}_{\mathrm{src}}(h_k)
    + \alpha_{\mathrm{src}}\,\mathbf{m}_k, \\
    \mathbf{z}^{(\ell)}_{\mathrm{tgt}}(t_k)
    &\leftarrow
    \mathbf{z}^{(\ell)}_{\mathrm{tgt}}(t_k)
    + \alpha_{\mathrm{tgt}}\,\mathbf{m}_k,
\end{align}
where $\alpha_{\mathrm{src}}, \alpha_{\mathrm{tgt}}\in\mathbb{R}$ are learnable scalars for source and target  endpoints, initialized to 1. We apply this injection before each layer, so the markers act as drag-specific encodings over the VAE tokens. Because the same marker $\mathbf{m}_k$ is added at both endpoints, the handle and target locations share a signature that biases the target location to attend to its handle (Appendix~\ref{app:attention}). Unlike training-free methods that impose correspondence at inference~\citep{yin2026lazydrag, he2025contextdrag}, DEM learns it during training.

\subsection{Joint Generation Modeling}

We represent an input image $I_{\mathrm{src}}$ and drag instruction $\mathcal{D}$ as an interleaved multimodal prompt $P$ with source-image tokens, textual drag tokens, and endpoint marker conditioning. The model supports direct editing, which generates the edited image from $P$, and reasoning-based editing, which first generates a reasoning chain $R=\{r_i\}_{i=1}^{M}$ and then conditions image generation on $(P,R)$.

For reasoning, we train the model with the standard autoregressive language-modeling objective
\begin{equation}
    \mathcal{L}_{\mathrm{think}}(\theta)
    =
    - \sum_{i=1}^{M}
    \log p_{\theta}(r_i \mid P, r_{<i}).
\end{equation}
The reasoning chain describes the source scene, interprets the semantic role of each drag, and summarizes the intended transformation. This supervision teaches the model to ground sparse point constraints in the visual content before image generation.

For image generation, we follow the rectified-flow objective~\citep{liuFlowStraightFast2023} used by BAGEL~\citep{dengEmergingPropertiesUnified2025}. Given the target latent $x_0$ and noise $x_1\sim\mathcal{N}(0,I)$, we sample $t\in[0,1]$ and define $x_t=(1-t)x_0+t x_1$ with target velocity $v=x_1-x_0$. The image-generation loss is
\begin{equation}
    \mathcal{L}_{\mathrm{img}}(\theta)
    =
    \mathbb{E}_{t,x_0,x_1}
    \left[
    \left\|
    v - v_{\theta}(x_t,t,P,R)
    \right\|_2^2
    \right],
\end{equation}
where $R$ is used for reasoning-based editing and omitted for direct editing. The final objective is $\mathcal{L}(\theta)=\mathcal{L}_{\mathrm{img}}(\theta)+\lambda_{\mathrm{think}}\mathcal{L}_{\mathrm{think}}(\theta)$,
where $\lambda_{\mathrm{think}}$ controls the weight of reasoning supervision.

\section{Experiments}
\label{sec:exp}

\subsection{Experimental Setup}

\paragraph{Benchmarks.}
We evaluate on DragBench~\citep{shi2024dragdiffusion}, the standard benchmark for drag-based image editing. We also introduce DragBench++, a new benchmark constructed using our dataset construction pipeline in Sec.~\ref{sec:dataset} for image pair generation. DragBench++ contains 145 samples with manually annotated drag instructions and emphasizes scenarios that require higher-level scene understanding, object-level reasoning, and semantic interpretation, with more drags per sample and longer displacements than DragBench (Table~\ref{tab:bench_stats}). Unlike prior benchmarks, each input also includes a pseudo-ground-truth edited image as reference. Although drag-based editing is under-constrained and multiple outputs may satisfy the same instruction, this reference edit verifies that each example has at least one plausible solution. It also enables reference-based evaluation and a more controlled analysis of whether existing metrics reflect meaningful editing quality. Reference edits are used only for evaluation. Baselines that require an edit mask receive manually annotated masks.

\paragraph{Evaluation Metrics.}
We evaluate along two axes: editing accuracy and generation quality. For editing accuracy, we follow prior work~\citep{shi2024dragdiffusion} and report \emph{Mean Distance} (MD), which measures the distance between each target point and the corresponding handle point tracked in the generated image using DIFT~\citep{tang2023emergent}, restricted to the edit mask. On DragBench++, reference edits show that DIFT-based correspondence can be unreliable under semantic ambiguity or large structural changes. We therefore also report MD$_{\text{MLLM}}$, which uses the same distance definition but replaces DIFT with GPT-5.5~\citep{gpt55} for point tracking, computed for the strongest method of each category (Appendix~\ref{app:mllm_eval}). Since GPT-5.5 is not used in our data pipeline, data construction and evaluation are decoupled. For generation quality, we report \emph{Image Fidelity} (IF) and FID~\citep{heusel2017gans}. Following prior work, IF is computed as $1-\mathrm{LPIPS}$~\citep{zhang2018unreasonable} between the input image and generated edit. Since fidelity to the input is ambiguous for editing tasks, where the output should preserve irrelevant content while changing the edited region, we also report FID on DragBench to measure overall visual quality and realism. For DragBench++, we instead report IF* against the reference edits as the primary fidelity metric.

\subsection{Comparison with Baselines}

\paragraph{Quantitative Evaluation.}
We compare against state-of-the-art drag-based image editing methods with public official implementations: test-time optimization methods DragDiffusion~\citep{shi2024dragdiffusion}, DragNoise~\citep{liu2024drag}, and DragLoRA~\citep{xiadraglora}; one-step methods FastDrag~\citep{zhao2024fastdrag} and GeoDrag~\citep{pu2026dragging}; and fine-tuned conditional generation methods InstantDrag~\citep{shin2024instantdrag} and LightningDrag~\citep{shi2025lightningdrag}. Table~\ref{tab:dragbench} reports quantitative results on DragBench and DragBench++, with \nickname\,in direct mode. Across both benchmarks, \nickname\,achieves the best generation quality, with the highest IF and lowest FID on DragBench, and the best IF and IF* on DragBench++. On DragBench, \nickname\,remains competitive in point-following accuracy, although the strongest MD scores come from test-time optimization and one-step methods that explicitly enforce the point displacement at inference.
The advantage is clearer on DragBench++, where edits require stronger semantic interpretation. \nickname\,achieves the best MD and MD$_{\text{MLLM}}$. The reference row also shows that DIFT-based MD is less reliable in this setting, since even reference edits obtain a relatively high MD. MD$_{\text{MLLM}}$ provides a more semantic estimate, under which \nickname\,substantially outperforms the baselines. The improvement in IF* further shows that \nickname\,does not improve point accuracy by producing implausible distortions; its outputs remain closest to the reference edits.

\setlength{\tabcolsep}{6pt}

\begin{table}[t]
\caption{Results on both benchmarks. TTO: test-time optimization, OS: one-step, FT: fine-tuned.}
\label{tab:dragbench}
\centering
\small
\resizebox{\linewidth}{!}{%
\begin{tabular}{l c c c c c c c c c}
\toprule
& & \multicolumn{4}{c}{\textsc{DragBench++}} & \multicolumn{4}{c}{\textsc{DragBench}} \\
\cmidrule(lr){3-6} \cmidrule(lr){7-10}
Method & Type
& IF~$\uparrow$ & IF*~$\uparrow$
& MD~$\downarrow$ & MD$_\text{MLLM}$~$\downarrow$
& IF~$\uparrow$ & FID~$\downarrow$ & MD~$\downarrow$ & MD$_\text{MLLM}$~$\downarrow$ \\
\midrule

Reference &
& -- & -- & 43.42 & 27.38
& -- & -- & -- & -- \\

\midrule

\textsc{DragDiffusion} & \multirow{3}{*}{\textsc{TTO}}
& 0.833 & 0.819 & 71.69 & --
& 0.882 & \underline{38.90} & 30.88 & -- \\

\textsc{DragNoise} &
& \underline{0.867} & 0.829 & 74.44 & --
& \underline{0.900} & 44.39 & 32.70 & -- \\

\textsc{DragLoRA} &
& 0.808 & 0.810 & \underline{48.36} & 59.98
& 0.874 & 43.35 & \underline{23.20} & \textbf{36.54} \\

\midrule

\textsc{FastDrag} & \multirow{2}{*}{\textsc{OS}}
& 0.780 & 0.771 & 79.02 & --
& 0.859 & 49.34 & 27.45 & -- \\

\textsc{GeoDrag} &
& 0.767 & 0.759 & 70.55 & 74.78
& 0.842 & 53.96 & \textbf{22.77} & 39.46 \\

\midrule

\textsc{InstantDrag} & \multirow{3}{*}{\textsc{FT}}
& 0.821 & 0.793 & 76.45 & --
& 0.882 & 45.10 & 30.78 & -- \\

\textsc{LightningDrag} &
& 0.842 & \underline{0.836} & 57.49 & \underline{57.28}
& 0.888 & 40.73 & 26.36 & 40.99 \\

\textbf{\nickname} &
& \textbf{0.875} & \textbf{0.905}
& \textbf{47.66} & \textbf{34.24}
& \textbf{0.918} & \textbf{30.05} & 28.84 & \underline{37.10} \\

\bottomrule
\end{tabular}%
\vspace{-10pt}
}
\end{table}

\begin{table}[t]
\centering
\small
\begin{minipage}[t]{0.49\linewidth}
  \centering
  \caption{Benchmark statistics.}
  \label{tab:bench_stats}
  \setlength{\tabcolsep}{2pt}
  \begin{tabular}{l c c}
  \toprule
  & \begin{tabular}[b]{@{}c@{}}\textsc{DragBench++}\\($n{=}145$)\end{tabular} & \begin{tabular}[b]{@{}c@{}}\textsc{DragBench}\\($n{=}205$)\end{tabular} \\
  \midrule
  Drags/sample      & 2.51  & 1.70  \\
  Multi-drag        & 80.7\% & 47.3\% \\
  Displacement (px)      & 107.8 & 58.5  \\
  Reference edit    & \checkmark & $\times$ \\
  \bottomrule
  \end{tabular}
\end{minipage}%
\hfill
\begin{minipage}[t]{0.49\linewidth}
  \centering
  \caption{User study: share of overall-preference votes (\%) on each benchmark.}
  \label{tab:user_study}
  \setlength{\tabcolsep}{2pt}
  \begin{tabular}{l c c}
  \toprule
  Method & \textsc{DragBench++} & \textsc{DragBench} \\
  \midrule
  \textsc{DragLoRA}      & 6.3 & 13.3 \\
  \textsc{GeoDrag}       & 6.3 & 8.7 \\
  \textsc{LightningDrag} & \underline{7.7} & \underline{17.0} \\
  \textbf{\nickname}     & \textbf{79.7} & \textbf{61.0} \\
  \bottomrule
  \end{tabular}
\end{minipage}
\vspace{-10pt}
\end{table}

\begin{figure}[t!]
\includegraphics[width=\linewidth]{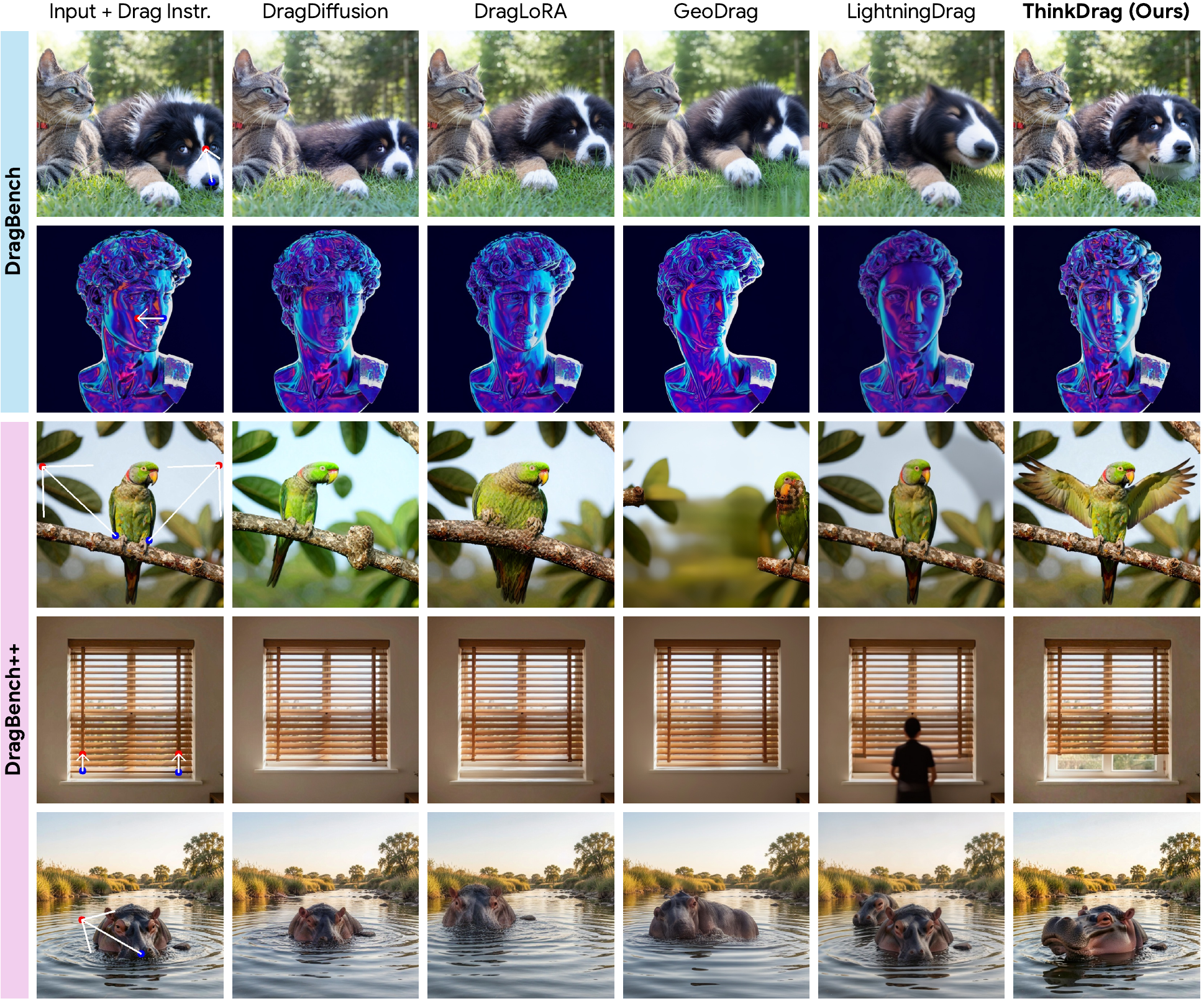}
    \caption{Qualitative comparison with other models on DragBench and DragBench++.}
    \vspace{-5pt}
    \label{fig:exp}
\end{figure}

\begin{figure}[t!]
\includegraphics[width=\linewidth]{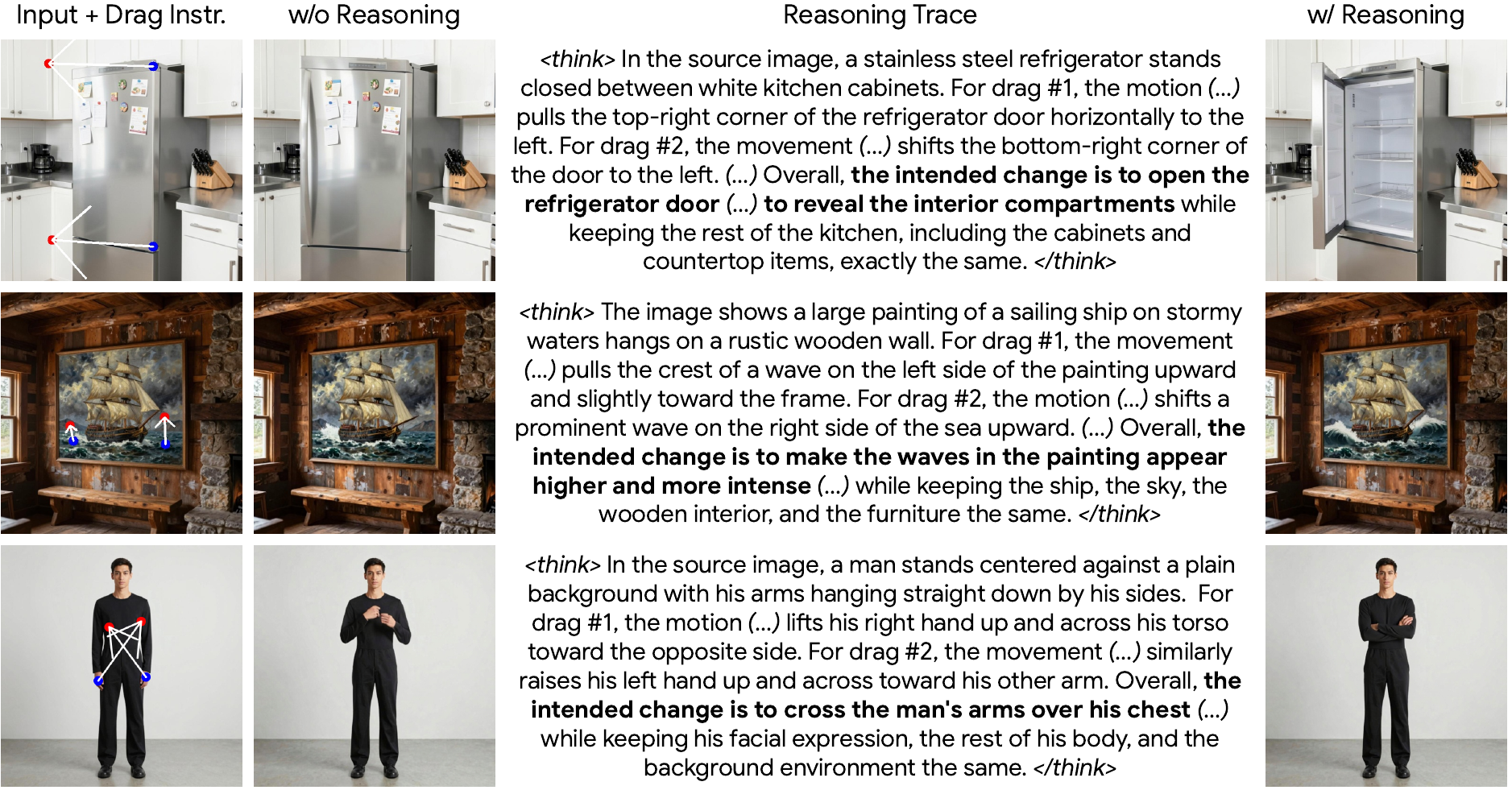}
    \caption{Qualitative comparison between direct generation and the reasoning mode on DragBench++.}
    \vspace{-10pt}
    \label{fig:think}
\end{figure}

\paragraph{Qualitative Analysis.}
Figure~\ref{fig:exp} shows qualitative comparisons on DragBench and DragBench++. On DragBench, most methods handle relatively direct point movements, while \nickname\,better preserves identity and local appearance during the edit. On DragBench++, the gap is more pronounced. Baselines often under-follow the drag instruction, introduce artifacts, or satisfy the point movement through implausible changes. In contrast, \nickname\,more consistently produces edits that are both spatially grounded and semantically plausible. Figure~\ref{fig:think} compares direct generation with the reasoning mode. Direct generation already benefits from semantic drag supervision, but reasoning can provide an additional advantage when the transformation is under-specified by the points alone. In such cases, explicitly interpreting the edit before generation helps resolve ambiguity and coordinate non-local changes, such as opening a refrigerator door or making waves taller in a painting.

\paragraph{User Study.}
We conduct a user study with 30 participants, each answering 20 questions sampled evenly from DragBench and DragBench++. For each question, participants see the input image, the drag instruction, and anonymized outputs of DragLoRA, GeoDrag, LightningDrag, and \nickname\,in random order, and select the preferred result for edit accuracy, generation quality, and overall preference. As shown in Table~\ref{tab:user_study}, participants clearly prefer \nickname, with 80\% of the overall-preference votes on DragBench++ and 61\% on DragBench, an external benchmark independent of our data pipeline, and similar shares for the other criteria (Appendix~\ref{app:user_study}). On DragBench, \nickname\,also receives 62\% of the edit-accuracy votes despite a higher DIFT-based MD than all three baselines, suggesting that MD does not fully capture semantic correctness.

\subsection{Joint Text-Drag Conditioning}

\begin{figure}[t]
\includegraphics[width=\linewidth]{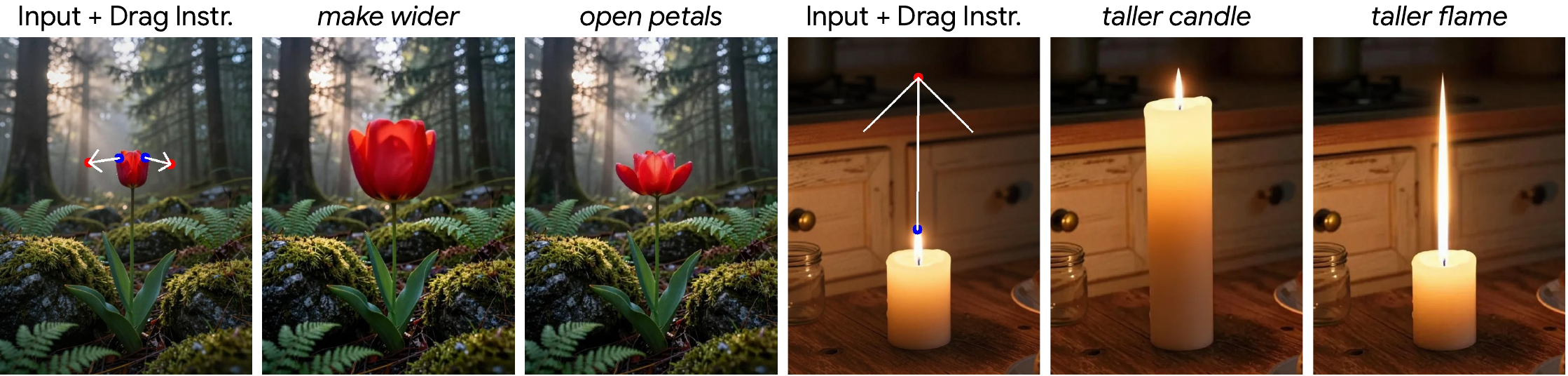}
    \caption{Joint text-drag conditioning. Given the same drag instruction, different text prompts guide the model toward different valid edits, while the drags keep each edit spatially grounded.}
    \label{fig:textdrag}
\end{figure}

In the default setting, \nickname\,infers the intended edit from the input image and drag points alone, either directly or through its reasoning chain. Joint text-drag conditioning provides a complementary mode where the user specifies part of the semantic intent. This is useful when the same point motion admits multiple plausible interpretations, since drag points provide spatial anchors while text clarifies the desired transformation. Prior work has explicitly targeted this setting with specialized mechanisms~\citep{jiangclipdrag}. In \nickname, it follows naturally from the unified multimodal formulation.
During training, we condition the model on both drag endpoints and the corresponding text edit description for 10\% of samples, allowing it to learn text-drag interaction without making text mandatory at inference time. As shown in Figure~\ref{fig:textdrag}, the same drag instruction can be resolved into different valid edits when paired with different text prompts, such as widening a flower versus opening its petals, or enlarging the candle body versus extending the flame.

\subsection{Ablation Studies}

\begin{figure}[t]
\centering
\begin{minipage}[t]{0.47\linewidth}
    \centering
    \captionof{table}{Reasoning and text conditioning on DragBench++ (MD$_{\text{MLLM}}$ with GPT-5.5).
    }
    \setlength{\tabcolsep}{6pt}
\small
\centering
\renewcommand{\arraystretch}{0.8}
\resizebox{\linewidth}{!}{%
\begin{tabular}{l c c}
\toprule
Method & IF*~$\uparrow$ & MD$_{\text{MLLM}}$~$\downarrow$ \\

\midrule

\nickname & 0.905 & 34.24 \\
w/ Reasoning & 0.911 & 33.92 \\
\midrule
w/ Reference Prompt & 0.914 & 33.82 \\

\bottomrule
\end{tabular}
}

    \label{tab:text_ablation}

    \captionof{table}{Component analysis on DragBench, from zero-shot base models to the full model.}
    \vspace{8pt}
    \setlength{\tabcolsep}{6pt}
\small
\centering
\resizebox{\linewidth}{!}{%
\begin{tabular}{l c c c}
\toprule
Method & IF~$\uparrow$ & FID~$\downarrow$ & MD~$\downarrow$ \\
\midrule
BAGEL (zero-shot) & 0.866 & 51.66 & 58.92 \\
FLUX.2 (zero-shot) & \textbf{0.933} & 32.30 & 58.56 \\
\midrule
\nickname & \underline{0.918} & \textbf{30.05} & \textbf{28.84} \\
w/o DEM & 0.911 & \underline{30.99} & \underline{45.42} \\
\bottomrule
\end{tabular}
}

    \label{tab:dem_ablation}
\end{minipage}
\hfill
\begin{minipage}[t]{0.49\linewidth}
    \vspace{-1mm}
    \centering
    \includegraphics[width=\linewidth]{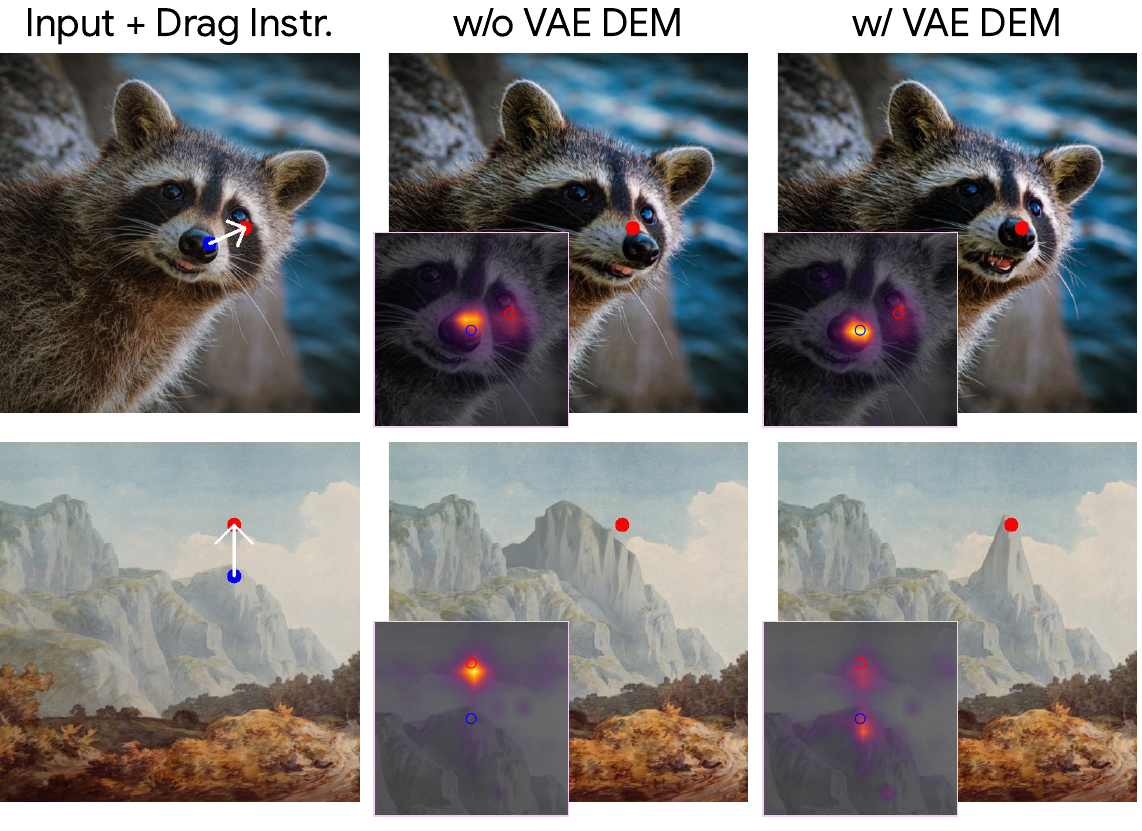}
    \captionof{figure}{Qualitative ablation of DEM. Insets show the averaged attention from the target point to the source image. With DEM, it peaks closer to the handle and the edit is more precise.}
    \label{fig:dem_ablation}
\end{minipage}
\vspace{-10pt}
\end{figure}

\paragraph{Reasoning and Language-based Conditioning.}
Table~\ref{tab:text_ablation} ablates reasoning and user-provided text conditioning. The reasoning mode improves reference-based fidelity over direct generation and slightly lowers MD$_{\text{MLLM}}$, approaching the setting in which the reference edit prompt removes the ambiguity of the drag instruction. The prompt itself changes MD$_{\text{MLLM}}$ only slightly, indicating that the drags already determine the spatial outcome, while the interpretation mainly affects how faithfully the intended edit is realized.
Per sample, reasoning yields the lower MD$_{\text{MLLM}}$ more often on multi-drag edits (56\%) than on single-drag edits (36\%; Appendix~\ref{app:reasoning}). Figure~\ref{fig:think} shows examples where reasoning helps resolve edits that require a more explicit interpretation of the intended transformation.

\paragraph{Drag Endpoint Markers.}
Table~\ref{tab:dem_ablation} separates the contributions of the base model, the data, and DEM. Neither BAGEL nor the editing model FLUX.2 follows drag instructions zero-shot, and FLUX.2 attains a high IF largely because it leaves the image unchanged. Training on our dataset without DEM substantially reduces MD, and adding DEM reduces it much further with similar IF and FID. This gap indicates that the structured text representation alone is insufficient for precise spatial control at this scale, and that endpoint markers provide important grounding. Training with textual drag coordinates nevertheless yields an emergent ability to ground coordinates in image content (Appendix~\ref{app:grounding}). Figure~\ref{fig:dem_ablation} further shows that without DEM the model still captures the overall intent of the drag instruction but lacks the spatial precision needed to align the edited content with the target point. The attention maps provide insight into this. Without DEM, the attention of the target location rarely peaks at the handle, resulting in the reduced precision observed. With DEM, the attention peaks at the handle for 51\% of drags on DragBench++, compared to 6\% without (Appendix~\ref{app:attention}).

\section{Conclusion}
\label{sec:conclusion}

We introduced \nickname, a unified multimodal framework for drag-based image editing with visual reasoning. By training on a new dataset of high-quality semantic drag transformations and reasoning traces, and by grounding each drag with Drag Endpoint Markers, \nickname\,learns to interpret sparse point constraints in relation to scene content and object structure, rather than treating them only as geometric displacements. We also introduced DragBench++, a benchmark of more challenging drag-based editing scenarios with reference edits. Experiments show that \nickname\,achieves state-of-the-art generation quality while maintaining competitive point-following precision, with the reasoning mode providing additional interpretability, particularly for ambiguous multi-drag edits.

\subsubsection*{Acknowledgments}
This research is supported by NTU SUG-NAP and is also supported by cash and in-kind funding from NTU S-Lab and industry partner(s).

{
    \small
    \bibliography{bib/arxiv}

@article{bai2023qwen,
  title = {Qwen Technical Report},
  author = {Bai, Jinze and Bai, Shuai and Chu, Yunfei and Cui, Zeyu and Dang, Kai and Deng, Xiaodong and Fan, Yang and Ge, Wenbin and Han, Yu and Huang, Fei and others},
  year = 2023,
  journal = {arXiv preprint arXiv:2309.16609},
  eprint = {2309.16609},
  archiveprefix = {arXiv}
}

@article{bai2025qwen3,
  title = {{Qwen3-VL} Technical Report},
  author = {Bai, Shuai and Cai, Yuxuan and Chen, Ruizhe and Chen, Keqin and Chen, Xionghui and Cheng, Zesen and Deng, Lianghao and Ding, Wei and Gao, Chang and Ge, Chunjiang and others},
  year = 2025,
  journal = {arXiv preprint arXiv:2511.21631},
  eprint = {2511.21631},
  archiveprefix = {arXiv}
}

@inproceedings{brooksInstructPix2PixLearningFollow2023,
  title = {{{InstructPix2Pix}}: {{Learning}} to {{Follow Image Editing Instructions}}},
  booktitle = {{CVPR}},
  author = {Brooks, Tim and Holynski, Aleksander and Efros, Alexei A.},
  year = 2023
}

@article{cai2025z,
  title = {Z-Image: {{An}} Efficient Image Generation Foundation Model with Single-Stream Diffusion Transformer},
  author = {Cai, Huanqia and Cao, Sihan and Du, Ruoyi and Gao, Peng and Hoi, Steven and Hou, Zhaohui and Huang, Shijie and Jiang, Dengyang and Jin, Xin and Li, Liangchen and others},
  year = 2025,
  journal = {arXiv preprint arXiv:2511.22699},
  eprint = {2511.22699},
  archiveprefix = {arXiv}
}

@article{chen2025janus,
  title = {Janus-pro: {{Unified}} Multimodal Understanding and Generation with Data and Model Scaling},
  author = {Chen, Xiaokang and Wu, Zhiyu and Liu, Xingchao and Pan, Zizheng and Liu, Wen and Xie, Zhenda and Yu, Xingkai and Ruan, Chong},
  year = 2025,
  journal = {arXiv preprint arXiv:2501.17811},
  eprint = {2501.17811},
  archiveprefix = {arXiv}
}

@inproceedings{choi2025dragtext,
  title = {Dragtext: {{Rethinking}} Text Embedding in Point-Based Image Editing},
  booktitle = {WACV},
  author = {Choi, Gayoon and Jeong, Taejin and Hong, Sujung and Hwang, Seong Jae},
  year = 2025,
}

@inproceedings{cui2024localize,
  title = {Localize, Understand, Collaborate: {{Semantic-aware}} Dragging via Intention Reasoner},
  author = {Cui, Xing and Li, Peipei and Li, Zekun and Liu, Xuannan and Zou, Yueying and He, Zhaofeng},
  year = 2024,
  booktitle = {NeurIPS},
}

@article{dengEmergingPropertiesUnified2025,
  title = {Emerging Properties in Unified Multimodal Pretraining},
  author = {Deng, Chaorui and Zhu, Deyao and Li, Kunchang and Gou, Chenhui and Li, Feng and Wang, Zeyu and Zhong, Shu and Yu, Weihao and Nie, Xiaonan and Song, Ziang and others},
  year = 2025,
  journal = {arXiv preprint arXiv:2505.14683},
  eprint = {2505.14683},
  archiveprefix = {arXiv}
}

@inproceedings{dhariwalDiffusionModelsBeat2021,
  title = {Diffusion {{Models Beat GANs}} on {{Image Synthesis}}},
  booktitle = {NeurIPS},
  author = {Dhariwal, Prafulla and Nichol, Alexander Quinn},
  year = 2021
}

@misc{flux2devturbo,
  title = {{{FLUX}}.2 [Dev] Turbo {{LoRA}}},
  author = {{fal}},
  year = 2025
}

@misc{flux2klein9b,
  title = {{{FLUX}}.2 [Klein] {{9B}}},
  author = {{Black Forest Labs}},
  year = 2026
}

@misc{gemini3,
  title = {Gemini 3 Developer Guide},
  author = {{Google}},
  year = 2025
}

@article{he2025contextdrag,
  title = {{{ContextDrag}}: {{Precise}} Drag-Based Image Editing via Context-Preserving Token Injection and Position-Consistent Attention},
  author = {He, Huiguo and Yan, Pengyu and Yi, Ziqi and Zhong, Weizhi and Liu, Zheng and Tang, Yejun and Yang, Huan and Gai, Kun and Li, Guanbin and Jin, Lianwen},
  year = 2025,
  journal = {arXiv preprint arXiv:2512.08477},
  eprint = {2512.08477},
  archiveprefix = {arXiv}
}

@article{he2026re,
  title = {Re-Align: {{Structured}} Reasoning-Guided Alignment for in-Context Image Generation and Editing},
  author = {He, Runze and Cheng, Yiji and Hang, Tiankai and Li, Zhimin and Xu, Yu and Yin, Zijin and Zhang, Shiyi and Dai, Wenxun and Du, Penghui and Ma, Ao and others},
  year = 2026,
  journal = {arXiv preprint arXiv:2601.05124},
  eprint = {2601.05124},
  archiveprefix = {arXiv}
}

@inproceedings{hertzPrompttoPromptImageEditing2023,
  title = {Prompt-to-{{Prompt Image Editing}} with {{Cross-Attention Control}}},
  booktitle = {ICLR},
  author = {Hertz, Amir and Mokady, Ron and Tenenbaum, Jay and Aberman, Kfir and Pritch, Yael and {Cohen-Or}, Daniel},
  year = 2023
}

@inproceedings{heusel2017gans,
  title = {Gans Trained by a Two Time-Scale Update Rule Converge to a Local Nash Equilibrium},
  author = {Heusel, Martin and Ramsauer, Hubert and Unterthiner, Thomas and Nessler, Bernhard and Hochreiter, Sepp},
  year = 2017,
  booktitle = {NeurIPS},
}

@inproceedings{hoDenoisingDiffusionProbabilistic2020,
  title = {Denoising {{Diffusion Probabilistic Models}}},
  booktitle = {NeurIPS},
  author = {Ho, Jonathan and Jain, Ajay and Abbeel, Pieter},
  year = 2020
}

@inproceedings{hou2024easydrag,
  title = {Easydrag: {{Efficient}} Point-Based Manipulation on Diffusion Models},
  booktitle = {CVPR},
  author = {Hou, Xingzhong and Liu, Boxiao and Zhang, Yi and Liu, Jihao and Liu, Yu and You, Haihang},
  year = 2024,
}

@inproceedings{huLoRALowRankAdaptation2022,
  title = {{{LoRA}}: {{Low-Rank Adaptation}} of {{Large Language Models}}},
  booktitle = {ICLR},
  author = {Hu, Edward J. and Shen, Yelong and Wallis, Phillip and {Allen-Zhu}, Zeyuan and Li, Yuanzhi and Wang, Shean and Wang, Lu and Chen, Weizhu},
  year = 2022
}

@inproceedings{Jhong_2026_WACV,
  title = {Dragonite: {{Single-step}} Drag-Based Image Editing with Geometric-Semantic Guidance},
  booktitle = {WACV},
  author = {Jhong, Meng-Ting and Huang, Tai-Ming and Chen, Shang-Fu and Cheng, Wen-Huang and Hua, Kai-Lung},
  year = 2026,
}

@inproceedings{jiangclipdrag,
  title = {{{CLIPDrag}}: {{Combining}} Text-Based and Drag-Based Instructions for Image Editing},
  booktitle = {ICLR},
  author = {Jiang, Ziqi and Wang, Zhen and Chen, Long},
  year = 2025
}

@article{labs2025flux,
  title = {{FLUX.1 Kontext}: {{Flow}} Matching for in-Context Image Generation and Editing in Latent Space},
  author = {{Black Forest Labs}},
  year = 2025,
  journal = {arXiv preprint arXiv:2506.15742},
  eprint = {2506.15742},
  archiveprefix = {arXiv}
}

@inproceedings{liang2025mixture,
  title = {Mixture-of-Transformers: A Sparse and Scalable Architecture for Multi-Modal Foundation Models},
  booktitle = {{{ICLR}} 2025 Workshop},
  author = {Liang, Weixin and YU, {\relax LILI} and Luo, Liang and Iyer, Srini and Dong, Ning and Zhou, Chunting and Ghosh, Gargi and Lewis, Mike and Yih, Wen-tau and Zettlemoyer, Luke and others},
  year = 2025
}

@article{liao2025mogao,
  title = {Mogao: {{An}} Omni Foundation Model for Interleaved Multi-Modal Generation},
  author = {Liao, Chao and Liu, Liyang and Wang, Xun and Luo, Zhengxiong and Zhang, Xinyu and Zhao, Wenliang and Wu, Jie and Li, Liang and Tian, Zhi and Huang, Weilin},
  year = 2025,
  journal = {arXiv preprint arXiv:2505.05472},
  eprint = {2505.05472},
  archiveprefix = {arXiv}
}

@inproceedings{lin2025gdragtowards,
  title = {{{GDrag}}: Towards General-Purpose Interactive Editing with Anti-Ambiguity Point Diffusion},
  booktitle = {ICLR},
  author = {Lin, Xiaojian and Li, Hanhui and Cheng, Yuhao and Yan, Yiqiang and Liang, Xiaodan},
  year = 2025
}

@inproceedings{ling2024freedrag,
  title = {Freedrag: {{Feature}} Dragging for Reliable Point-Based Image Editing},
  booktitle = {CVPR},
  author = {Ling, Pengyang and Chen, Lin and Zhang, Pan and Chen, Huaian and Jin, Yi and Zheng, Jinjin},
  year = 2024,
}

@inproceedings{liu2024drag,
  title = {Drag Your Noise: {{Interactive}} Point-Based Editing via Diffusion Semantic Propagation},
  booktitle = {CVPR},
  author = {Liu, Haofeng and Xu, Chenshu and Yang, Yifei and Zeng, Lihua and He, Shengfeng},
  year = 2024,
}

@inproceedings{liu2024improved,
  title = {Improved Baselines with Visual Instruction Tuning},
  booktitle = {CVPR},
  author = {Liu, Haotian and Li, Chunyuan and Li, Yuheng and Lee, Yong Jae},
  year = 2024,
}

@inproceedings{liuFlowStraightFast2023,
  title = {Flow Straight and Fast: {{Learning}} to Generate and Transfer Data with Rectified Flow},
  booktitle = {ICLR},
  author = {Liu, Xingchao and Gong, Chengyue and others},
  year = 2023
}

@inproceedings{liuVisualInstructionTuning2023,
  title = {Visual {{Instruction Tuning}}},
  booktitle = {{NeurIPS}},
  author = {Liu, Haotian and Li, Chunyuan and Wu, Qingyang and Lee, Yong Jae},
  year = 2023
}

@inproceedings{lu2024regiondrag,
  title = {Regiondrag: {{Fast}} Region-Based Image Editing with Diffusion Models},
  booktitle = {ECCV},
  author = {Lu, Jingyi and Li, Xinghui and Han, Kai},
  year = 2024,
}

@inproceedings{lu2025inpaint4drag,
  title = {{{Inpaint4Drag}}: {{Repurposing}} Inpainting Models for Drag-Based Image Editing via Bidirectional Warping},
  booktitle = {ICCV},
  author = {Lu, Jingyi and Han, Kai},
  year = 2025,
}

@inproceedings{meng2022sdedit,
  title = {{{SDEdit}}: {{Guided}} Image Synthesis and Editing with Stochastic Differential Equations},
  booktitle = {ICLR},
  author = {Meng, Chenlin and He, Yutong and Song, Yang and Song, Jiaming and Wu, Jiajun and Zhu, Jun-Yan and Ermon, Stefano},
  year = 2022
}

@inproceedings{moudragondiffusion,
  title = {{{DragonDiffusion}}: {{Enabling}} Drag-Style Manipulation on Diffusion Models},
  booktitle = {ICLR},
  author = {Mou, Chong and Wang, Xintao and Song, Jiechong and Shan, Ying and Zhang, Jian},
  year = 2024
}

@inproceedings{nieblessing,
  title = {The Blessing of Randomness: {{SDE}} Beats {{ODE}} in General Diffusion-Based Image Editing},
  booktitle = {ICLR},
  author = {Nie, Shen and Guo, Hanzhong Allan and Lu, Cheng and Zhou, Yuhao and Zheng, Chenyu and Li, Chongxuan},
  year = 2024
}

@inproceedings{panDragYourGan2023,
  title = {Drag Your {GAN}: {{Interactive}} Point-Based Manipulation on the Generative Image Manifold},
  booktitle = {ACM SIGGRAPH},
  author = {Pan, Xingang and Tewari, Ayush and Leimk{\"u}hler, Thomas and Liu, Lingjie and Meka, Abhimitra and Theobalt, Christian},
  year = 2023,
}

@inproceedings{pu2026dragging,
  title = {Dragging with Geometry: {{From}} Pixels to Geometry-Guided Image Editing},
  booktitle = {ICLR},
  author = {Pu, Xinyu and Wang, Hongsong and Gui, Jie and Zhou, Pan},
  year = 2026
}

@inproceedings{qin2026unicot,
  title = {Uni-{{CoT}}: {{Towards}} Unified Chain-of-Thought Reasoning across Text and Vision},
  booktitle = {ICLR},
  author = {Qin, Luozheng and JIA, {\relax GONG} and Sun, Yuqing and Li, Tianjiao and Pan, Haoyu and Yang, Mengping and Yang, Xiaomeng and Qu, Chao and Tan, Zhiyu and Li, Hao},
  year = 2026
}

@inproceedings{shao2024visual,
  title = {Visual Cot: {{Advancing}} Multi-Modal Language Models with a Comprehensive Dataset and Benchmark for Chain-of-Thought Reasoning},
  author = {Shao, Hao and Qian, Shengju and Xiao, Han and Song, Guanglu and Zong, Zhuofan and Wang, Letian and Liu, Yu and Li, Hongsheng},
  year = 2024,
  booktitle = {NeurIPS},
}

@inproceedings{shi2024dragdiffusion,
  title = {Dragdiffusion: {{Harnessing}} Diffusion Models for Interactive Point-Based Image Editing},
  booktitle = {CVPR},
  author = {Shi, Yujun and Xue, Chuhui and Liew, Jun Hao and Pan, Jiachun and Yan, Hanshu and Zhang, Wenqing and Tan, Vincent YF and Bai, Song},
  year = 2024,
}

@inproceedings{shi2025lightningdrag,
  title = {{{LightningDrag}}: {{Lightning}} Fast and Accurate Drag-Based Image Editing Emerging from Videos},
  booktitle = {ICML},
  author = {Shi, Yujun and Liew, Jun Hao and Yan, Hanshu and Tan, Vincent Y. F. and Feng, Jiashi},
  year = 2025
}

@inproceedings{shin2024instantdrag,
  title = {Instantdrag: {{Improving}} Interactivity in Drag-Based Image Editing},
  booktitle = {ACM SIGGRAPH Asia},
  author = {Shin, Joonghyuk and Choi, Daehyeon and Park, Jaesik},
  year = 2024,
}

@article{simeoniDINOv32025,
  title = {{{DINOv3}}},
  author = {Sim{\'e}oni, Oriane and Vo, Huy V. and Seitzer, Maximilian and Baldassarre, Federico and Oquab, Maxime and Jose, Cijo and Khalidov, Vasil and Szafraniec, Marc and Yi, Seungeun and Ramamonjisoa, Micha{\"e}l and Massa, Francisco and Haziza, Daniel and Wehrstedt, Luca and Wang, Jianyuan and Darcet, Timoth{\'e}e and Moutakanni, Th{\'e}o and Sentana, Leonel and Roberts, Claire and Vedaldi, Andrea and Tolan, Jamie and Brandt, John and Couprie, Camille and Mairal, Julien and J{\'e}gou, Herv{\'e} and Labatut, Patrick and Bojanowski, Piotr},
  year = 2025,
  journal = {arXiv preprint arXiv:2508.10104},
  eprint = {2508.10104},
}

@inproceedings{songDenoisingDiffusionImplicit2021,
  title = {Denoising {{Diffusion Implicit Models}}},
  booktitle = {ICLR},
  author = {Song, Jiaming and Meng, Chenlin and Ermon, Stefano},
  year = 2021,
}

@inproceedings{tang2023emergent,
  title = {Emergent Correspondence from Image Diffusion},
  author = {Tang, Luming and Jia, Menglin and Wang, Qianqian and Phoo, Cheng Perng and Hariharan, Bharath},
  year = 2023,
  booktitle = {NeurIPS},
}

@article{team2024chameleon,
  title = {Chameleon: {{Mixed-modal}} Early-Fusion Foundation Models},
  author = {{Chameleon Team}},
  year = 2024,
  journal = {arXiv preprint arXiv:2405.09818},
  eprint = {2405.09818},
  archiveprefix = {arXiv}
}

@article{wang2024emu3,
  title = {Emu3: {{Next-token}} Prediction Is All You Need},
  author = {Wang, Xinlong and Zhang, Xiaosong and Luo, Zhengxiong and Sun, Quan and Cui, Yufeng and Wang, Jinsheng and Zhang, Fan and Wang, Yueze and Li, Zhen and Yu, Qiying and others},
  year = 2024,
  journal = {arXiv preprint arXiv:2409.18869},
  eprint = {2409.18869},
  archiveprefix = {arXiv}
}

@article{wang2025multimodal,
  title = {Multimodal Chain-of-Thought Reasoning: {{A}} Comprehensive Survey},
  author = {Wang, Yaoting and Wu, Shengqiong and Zhang, Yuecheng and Yan, Shuicheng and Liu, Ziwei and Luo, Jiebo and Fei, Hao},
  year = 2025,
  journal = {arXiv preprint arXiv:2503.12605},
  eprint = {2503.12605},
  archiveprefix = {arXiv}
}

@inproceedings{wu2025janus,
  title = {Janus: {{Decoupling}} Visual Encoding for Unified Multimodal Understanding and Generation},
  booktitle = {CVPR},
  author = {Wu, Chengyue and Chen, Xiaokang and Wu, Zhiyu and Ma, Yiyang and Liu, Xingchao and Pan, Zizheng and Liu, Wen and Xie, Zhenda and Yu, Xingkai and Ruan, Chong and others},
  year = 2025,
}

@article{wu2025qwen,
  title = {Qwen-Image Technical Report},
  author = {Wu, Chenfei and Li, Jiahao and Zhou, Jingren and Lin, Junyang and Gao, Kaiyuan and Yan, Kun and Yin, Sheng-ming and Bai, Shuai and Xu, Xiao and Chen, Yilei and others},
  year = 2025,
  journal = {arXiv preprint arXiv:2508.02324},
  eprint = {2508.02324},
  archiveprefix = {arXiv}
}

@inproceedings{xiadraglora,
  title = {{{DragLoRA}}: {{Online}} Optimization of {{LoRA}} Adapters for Drag-Based Image Editing in Diffusion Model},
  booktitle = {ICML},
  author = {Xia, Siwei and Sun, Li and Sun, Tiantian and Li, Qingli},
  year = 2025
}

@article{xie2025show,
  title = {Show-{{O2}}: {{Improved}} Native Unified Multimodal Models},
  author = {Xie, Jinheng and Yang, Zhenheng and Shou, Mike Zheng},
  year = 2025,
  journal = {arXiv preprint arXiv:2506.15564},
  eprint = {2506.15564},
  archiveprefix = {arXiv}
}

@inproceedings{xieshow,
  title = {Show-o: {{One}} Single Transformer to Unify Multimodal Understanding and Generation},
  booktitle = {ICLR},
  author = {Xie, Jinheng and Mao, Weijia and Bai, Zechen and Zhang, David Junhao and Wang, Weihao and Lin, Kevin Qinghong and Gu, Yuchao and Chen, Zhijie and Yang, Zhenheng and Shou, Mike Zheng},
  year = 2025
}

@inproceedings{yin2026lazydrag,
  title = {{{LazyDrag}}: {{Enabling}} Stable Drag-Based Editing on Multi-Modal Diffusion Transformers via Explicit Correspondence},
  booktitle = {ICLR},
  author = {Yin, Zixin and Dai, Xili and Wang, Duomin and Zeng, Xianfang and Ni, Lionel and YU, Gang and Shum, Heung-Yeung},
  year = 2026
}

@inproceedings{zhang2018unreasonable,
  title = {The Unreasonable Effectiveness of Deep Features as a Perceptual Metric},
  booktitle = {CVPR},
  author = {Zhang, Richard and Isola, Phillip and Efros, Alexei A and Shechtman, Eli and Wang, Oliver},
  year = 2018,
}

@inproceedings{zhanggooddrag,
  title = {{{GoodDrag}}: {{Towards}} Good Practices for Drag Editing with Diffusion Models},
  booktitle = {ICLR},
  author = {Zhang, Zewei and Liu, Huan and Chen, Jun and Xu, Xiangyu},
  year = 2025
}

@inproceedings{zhao2024fastdrag,
  title = {{{FastDrag}}: {{Manipulate}} Anything in One Step},
  booktitle = {NeurIPS},
  author = {Zhao, Xuanjia and Guan, Jian and Fan, Congyi and Xu, Dongli and Lin, Youtian and Pan, Haiwei and Feng, Pengming},
  year = 2024
}

@inproceedings{zhou2026dragnext,
  title = {Dragnext: {{Rethinking}} Drag-Based Image Editing},
  booktitle = {AAAI},
  author = {Zhou, Yuan and Zhou, Junbao and Xu, Qingshan and Zhao, Kesen and Wang, Yuxuan and Fei, Hao and Hong, Richang and Zhang, Hanwang},
  year = 2026,
}

@inproceedings{zhoutransfusion,
  title = {Transfusion: {{Predict}} the next Token and Diffuse Images with One Multi-Modal Model},
  booktitle = {ICLR},
  author = {Zhou, Chunting and YU, {\relax LILI} and Babu, Arun and Tirumala, Kushal and Yasunaga, Michihiro and Shamis, Leonid and Kahn, Jacob and Ma, Xuezhe and Zettlemoyer, Luke and Levy, Omer},
  year = 2025
}

@misc{gpt55,
  title  = {{GPT-5.5} System Card},
  author = {{OpenAI}},
  year   = {2026},
  month  = apr,
  howpublished = {\url{https://openai.com/index/gpt-5-5-system-card}}
}
    \bibliographystyle{iclr2027_conference}
}

\clearpage
\appendix
\section{Additional Experiments}

\subsection{Additional Visual Results}
\label{app:additional_visual}

\begin{figure}[p]
\includegraphics[width=\linewidth]{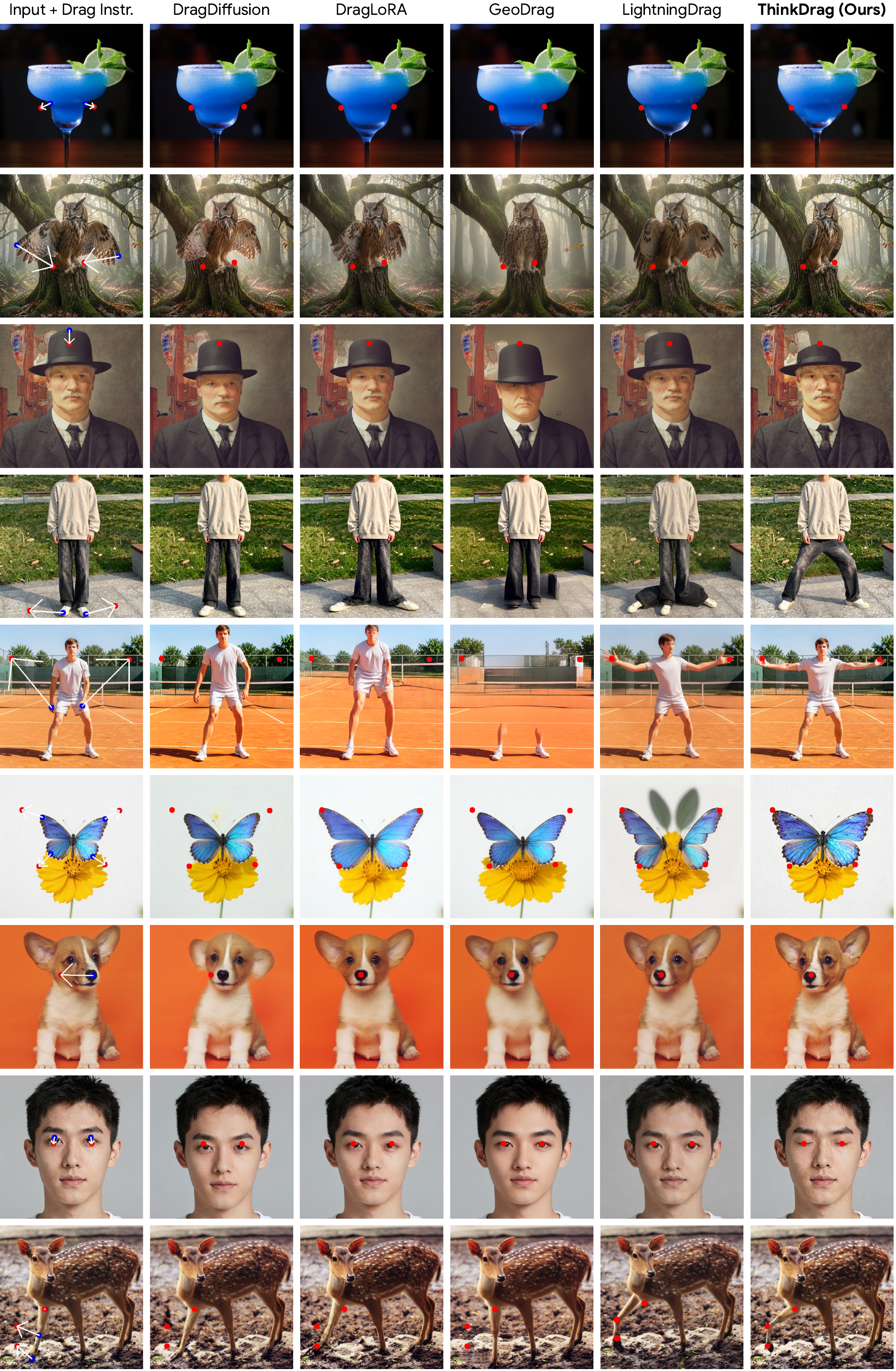}
    \caption{Additional visual comparison with other models on DragBench and DragBench++. Target points are overlaid on the generated images to facilitate analysis of editing accuracy.}
    \label{fig:exp_add}
\end{figure}

Figure~\ref{fig:exp_add} provides additional qualitative comparisons on DragBench and DragBench++. For easier inspection of editing accuracy, we overlay the target points on the generated images. These examples further illustrate the ability of \nickname\,to produce spatially accurate edits while preserving image identity and visual quality.

\subsection{Reasoning Analysis}
\label{app:reasoning}

Table~\ref{tab:winrate} reports the win rate of reasoning over direct generation on DragBench++, where a mode wins a sample if its output has the lower MD$_{\text{MLLM}}$. Overall, the two modes win about equally often (reasoning wins 75 of the 145 samples, 51.7\%), but reasoning wins on 56\% of multi-drag edits compared with 36\% of single-drag edits. Table~\ref{tab:winrate_type} shows win rates across edit types. To assess faithfulness, Qwen3.6-27B judges whether each generated trace describes the same edit as the annotated edit prompt, and as a control we construct a deliberately incorrect edit prompt for each sample. The traces are judged consistent with the annotated prompt for 78.6\% of samples and with the incorrect prompt for 2.8\%. Since an ambiguous drag can admit several valid interpretations, the first number is a conservative estimate of trace correctness.

\begin{table}[h]
\centering
\small
\caption{Win rate of reasoning over direct generation on DragBench++ by drag number (MD$_{\text{MLLM}}$).}
\label{tab:winrate}
\begin{tabular}{l c c c c c}
\toprule
Number of drags & 1 & 2 & 3 & $\geq$4 & All \\
\midrule
Samples & 28 & 51 & 37 & 29 & 145 \\
Reasoning wins & 10 & 25 & 24 & 16 & 75 \\
Win rate & 35.7\% & 49.0\% & 64.9\% & 55.2\% & 51.7\% \\
\bottomrule
\end{tabular}
\end{table}

\begin{table}[h]
\centering
\small
\caption{Win rate of reasoning over direct generation on DragBench++ by edit type, determined by the first verb of the edit instruction. Verbs with fewer than five samples are grouped under Other.}
\label{tab:winrate_type}
\begin{tabular}{l c c c c c c}
\toprule
Edit type & Open & Adjust & Move & Rotate & Raise & Other \\
\midrule
Samples & 20 & 23 & 20 & 19 & 19 & 44 \\
Win rate & 60.0\% & 52.2\% & 50.0\% & 47.4\% & 42.1\% & 54.5\% \\
\bottomrule
\end{tabular}
\end{table}

\subsection{Coordinate Grounding}
\label{app:grounding}

\begin{figure}[t]
\centering
\includegraphics[width=\linewidth]{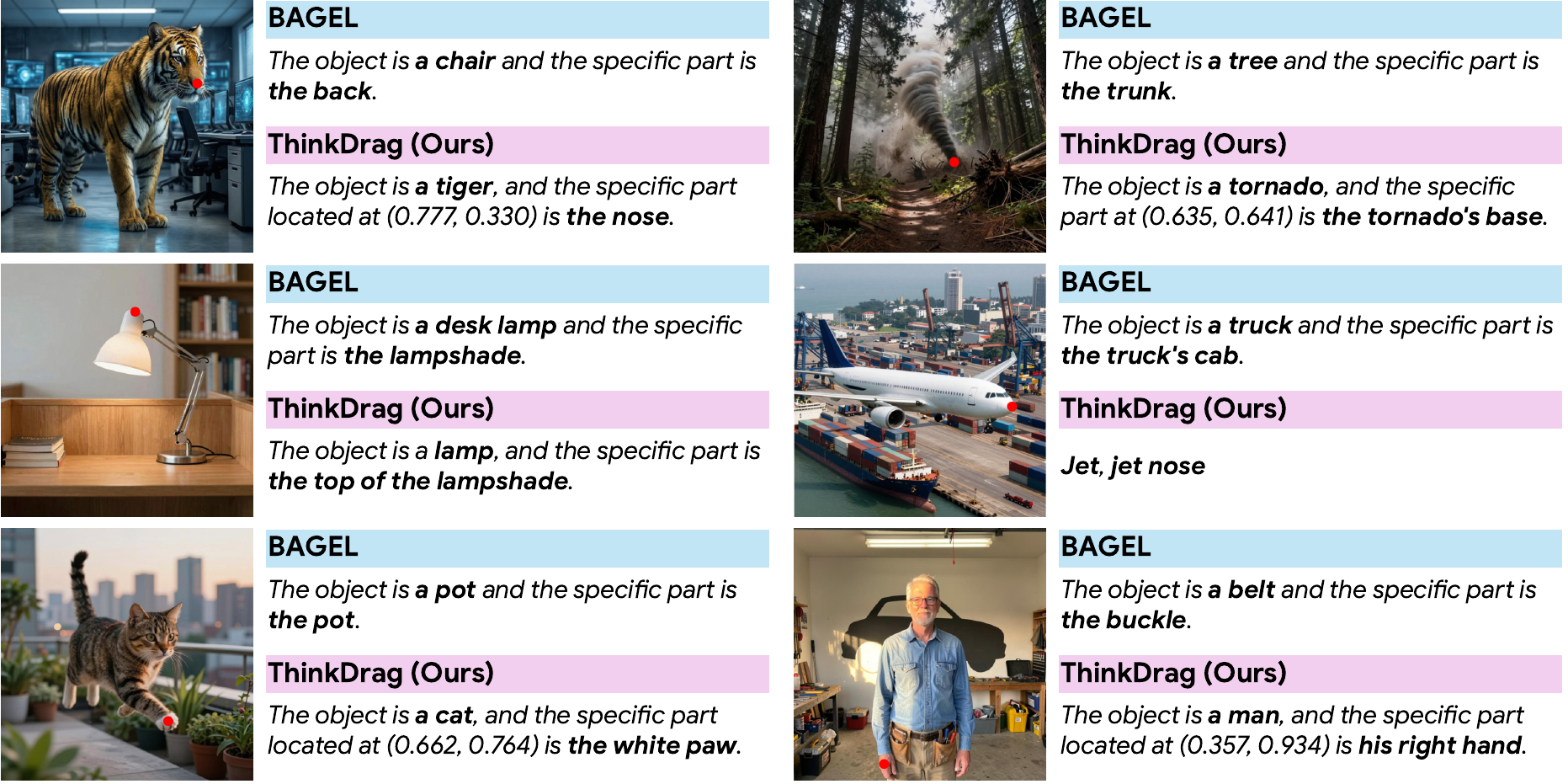}
\caption{Grounding of textual coordinates. Each input image shows the queried normalized coordinate as a red point. Given the image and the coordinate, BAGEL often describes an unrelated object, whereas \nickname\,identifies the correct object and part at the queried location.}
\label{fig:grounding}
\end{figure}

Training with textual drag coordinates gives rise to an emergent ability to ground coordinates in image content. To evaluate this ability, we manually annotate the object and object part at 250 handle points from DragBench++. Given the image and a normalized coordinate, each model is asked to identify the object and the object part at that position, and Qwen3.6-27B compares the answers with the labels. Table~\ref{tab:grounding} reports the results. Figure~\ref{fig:grounding} shows examples where \nickname\,identifies the correct object and part at the queried coordinate, while BAGEL describes an unrelated object.

\begin{table}[h]
\centering
\small
\caption{Coordinate grounding on 250 annotated handle points from DragBench++.}
\label{tab:grounding}
\begin{tabular}{l c c}
\toprule
Model & Correct & Accuracy \\
\midrule
BAGEL & 95 & 38.0\% \\
\nickname & 182 & 72.8\% \\
\bottomrule
\end{tabular}
\end{table}

\subsection{Attention Analysis: Target Point to Handle Correspondence}
\label{app:attention}

We process every DragBench and DragBench++ sample with and without DEM from the same initial noise. For each drag, we extract the attention from the generated-image token at $t_k$ to the source VAE tokens, averaged over all heads, layers, and denoising steps. Since each region of the generated image draws its content predominantly from the source region it attends to most, the location of this attention peak is a good indicator of which source content matters most for the transfer to the target point. A drag executed as intended should therefore have its attention peak at the corresponding handle. A drag counts as a hit if the attention peak lies in the latent cell of $h_k$ or one of its eight neighbors, where one cell corresponds to 16 pixels. Drags shorter than three cells are excluded because their handle and target patches overlap. Table~\ref{tab:attn} summarizes the results. DEM substantially increases the fraction of drags whose attention peaks at the handle on both benchmarks.

\begin{table}[h]
\centering
\small
\caption{Fraction of drags whose cross-image attention from the target point peaks at the handle (within one latent cell), with and without DEM.}
\label{tab:attn}
\begin{tabular}{l c c c}
\toprule
Benchmark & Drags & w/o DEM & w/ DEM \\
\midrule
\textsc{DragBench++} & 294 & 6.1\% & 50.7\% \\
\textsc{DragBench} & 226 & 13.3\% & 46.0\% \\
\bottomrule
\end{tabular}
\end{table}

\subsection{Runtime and Memory Measurement}

We report runtime and peak GPU memory usage in Table~\ref{tab:runtime_memory}, measured on a single NVIDIA H200 GPU. The reasoning mode increases runtime due to autoregressive reasoning-chain generation before image synthesis, while peak memory remains nearly unchanged.

\begin{table}[h]
\caption{Runtime and peak memory usage.}
\label{tab:runtime_memory}
\centering
\small
\setlength{\tabcolsep}{6pt}
\begin{tabular}{l c c}
\toprule
Method & Runtime (s) & Peak Memory (GB) \\
\midrule
\nickname & 14.3 & 35.5 \\
\nickname\, w/ Reasoning & 22.9 & 35.6 \\
\bottomrule
\end{tabular}
\end{table}

\section{Additional Experimental Details}

\subsection{Implementation Details}\label{app:impl}

We implement \nickname\,using BAGEL~\citep{dengEmergingPropertiesUnified2025} as the base unified multimodal model. We train LoRA adapters~\citep{huLoRALowRankAdaptation2022} of rank 256 on the attention and MLP projections of the base transformer for 100K steps, using AdamW with a learning rate of $2\times10^{-5}$ and a cosine schedule. The effective batch size is approximately 16 samples, distributed across 4 NVIDIA H200 GPUs. The training procedure takes around 60 hours. We set the reasoning loss weight to $\lambda_{\mathrm{think}}=0.25$. We support up to 10 drag pairs per image during training; this limit is configurable. Unless otherwise specified, inference uses 50 sampling steps with classifier-free guidance scales of 3.0 for text and 1.5 for image conditioning. The learned endpoint scales remain close to their initialization ($\alpha_{\mathrm{tgt}}\approx1.16$, $\alpha_{\mathrm{src}}\approx1.12$). Images are processed at their native aspect ratio, with training images mostly square and the remainder in 4:3 and 16:9 in both orientations. The number of drags per sample (1--7) follows the annotation without random subsampling. The arrows and points drawn in our figures serve only as visualization, since the model receives drags solely as text and DEM. For the zero-shot rows of Table~\ref{tab:dem_ablation}, BAGEL and FLUX.2 receive the drags as structured text instructions.

\subsection{MLLM-based Evaluation}
\label{app:mllm_eval}

\begin{figure}[t]
\centering
\includegraphics[width=\linewidth]{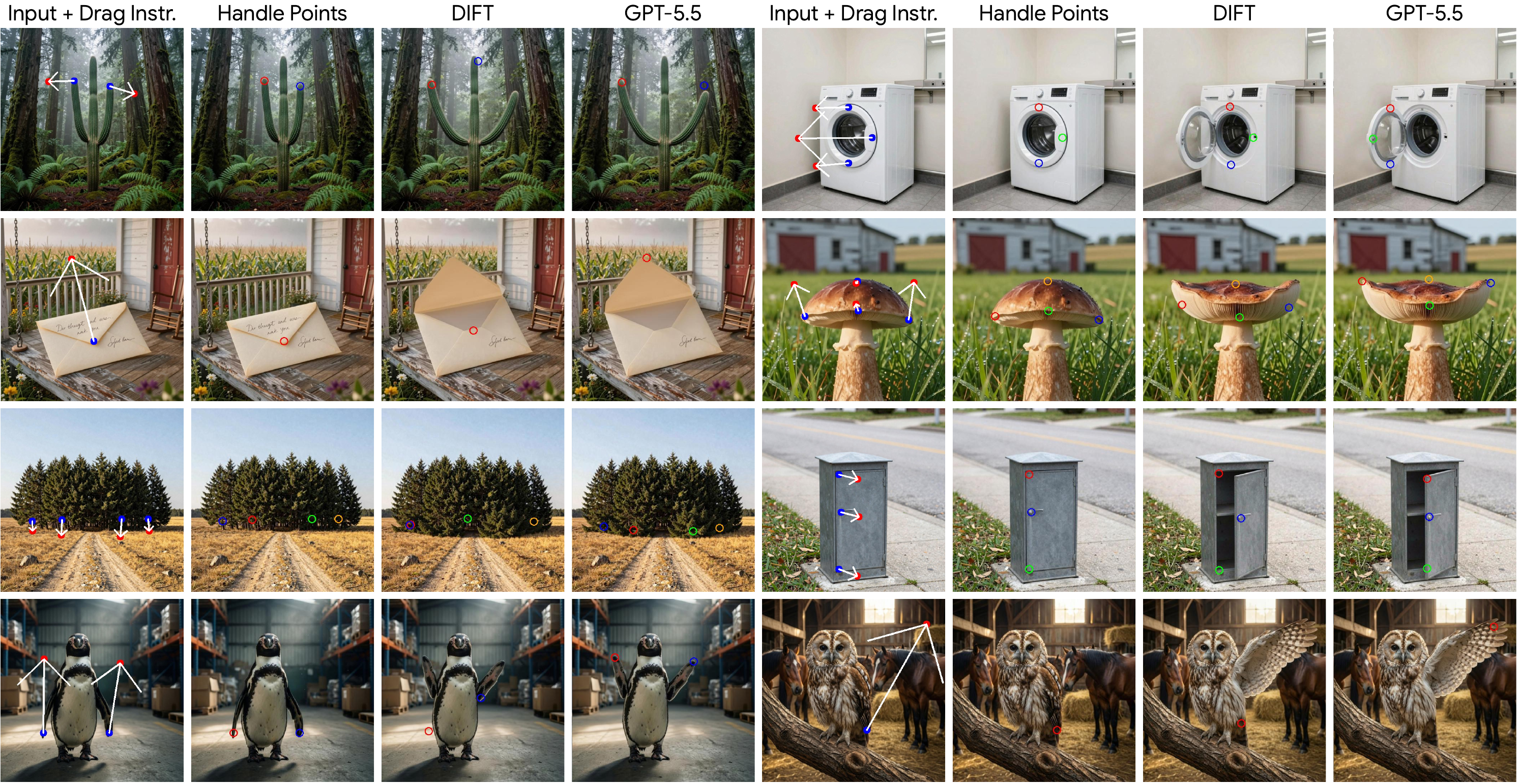}
\caption{Comparison between DIFT-based and MLLM-based point tracking. MLLM-based tracking is more robust in challenging cases involving large structural changes or distractors.}
\label{fig:mllm_eval}
\end{figure}

For the MD$_{\text{MLLM}}$ metric, we use GPT-5.5~\citep{gpt55} to track source handle points in the generated image. The model receives the source image, the generated image, and the source handle-point coordinates. It is prompted to return the corresponding physical or semantic point locations in the generated image using normalized coordinates in $[0,1000]$. Due to API cost, we compute MD$_{\text{MLLM}}$ for the strongest method of each category (DragLoRA, GeoDrag, and LightningDrag), the same baselines as in the user study, and mark the remaining methods with ``--'' in Table~\ref{tab:dragbench}. Figure~\ref{fig:mllm_eval} compares this protocol with DIFT-based tracking. Although MLLM-based tracking is also imperfect, it is more robust in challenging cases involving semantic ambiguity, large structural changes, or visually similar distractors, where local feature matching can select an incorrect correspondence.

On the reference edits of DragBench++, which satisfy their drags by construction, DIFT reports a mean distance of 43.42, whereas GPT-5.5 reports 27.38. Most of the DIFT distance on this benchmark therefore reflects tracking failures under large structural changes. On DragBench, where edits are more local, DIFT remains accurate, so we report both trackers and regard them as complementary.

\begin{tcolorbox}[
    colback=gray!5,
    colframe=gray!50,
    title={Prompt for MLLM-based point tracking},
    fonttitle=\bfseries,
    breakable
]
\small
\begin{verbatim}
You are a highly precise vision-language model. You see two images in order:
SOURCE then GENERATED.

Your task is point tracking. The user provides handle point coordinates in the
SOURCE image. For each provided source handle point, return the corresponding
same physical/semantic point location in the GENERATED image.

Critical rules:
- Track the exact same point or landmark.
- The source handle point may have moved, deformed, or changed appearance in
  the generated image.
- Use nearby object geometry, part boundaries, corners/tips/joints/hinges,
  texture, and semantic context to identify the correspondence.
- If a point is occluded or not visible in the GENERATED image, set its target
  to the nearest visible point.
- Coordinates are normalized to 0-1000: (0,0) top-left, (1000,1000)
  bottom-right of the GENERATED image.
- Return exactly one entry per input point, preserving the input point index.
- Do not include markdown or prose.

Output STRICT JSON:
{"tracked_points":[{"index":0,"target":[x,y]}]}
\end{verbatim}
\end{tcolorbox}

\subsection{User Study}
\label{app:user_study}

Each of the 30 participants answered 20 questions sampled evenly from DragBench and DragBench++. For each question, edited results from different methods were anonymized and shown in random order. \emph{Edit accuracy} measures how well the output follows the drag instruction and moves the intended content toward the target location. \emph{Generation quality} measures whether the edited image remains faithful to the input, preserving identity, appearance, and unrelated regions while avoiding artifacts. \emph{Overall preference} considers both criteria. We compare against DragLoRA, GeoDrag, and LightningDrag, the strongest baseline of each category. Table~\ref{tab:user_study_full} reports all three criteria for each benchmark and for both benchmarks combined. Over both benchmarks, \nickname\,receives about 70\% of the votes under each criterion and the most votes on 18 of the 20 questions.

\begin{table}[h]
\centering
\small
\caption{User study vote share (\%) by benchmark and criterion. Each benchmark contributes 10 of the 20 questions.}
\label{tab:user_study_full}
\begin{tabular}{l l c c c}
\toprule
Benchmark & Method & Edit accuracy & Generation quality & Overall preference \\
\midrule
\multirow{4}{*}{\textsc{DragBench++}} & \textsc{DragLoRA} & \underline{7.7} & 6.3 & 6.3 \\
 & \textsc{GeoDrag} & 7.0 & 6.7 & 6.3 \\
 & \textsc{LightningDrag} & \underline{7.7} & \underline{8.3} & \underline{7.7} \\
 & \textbf{\nickname} & \textbf{77.7} & \textbf{78.7} & \textbf{79.7} \\
\midrule
\multirow{4}{*}{\textsc{DragBench}} & \textsc{DragLoRA} & 13.7 & \underline{15.7} & 13.3 \\
 & \textsc{GeoDrag} & 7.7 & 8.0 & 8.7 \\
 & \textsc{LightningDrag} & \underline{16.7} & \underline{15.7} & \underline{17.0} \\
 & \textbf{\nickname} & \textbf{62.0} & \textbf{60.7} & \textbf{61.0} \\
\midrule
\multirow{4}{*}{Both} & \textsc{DragLoRA} & 10.7 & 11.0 & 9.8 \\
 & \textsc{GeoDrag} & 7.3 & 7.3 & 7.5 \\
 & \textsc{LightningDrag} & \underline{12.2} & \underline{12.0} & \underline{12.3} \\
 & \textbf{\nickname} & \textbf{69.8} & \textbf{69.7} & \textbf{70.3} \\
\bottomrule
\end{tabular}
\end{table}

\begin{figure}[t]
\centering
\includegraphics[width=0.6\linewidth]{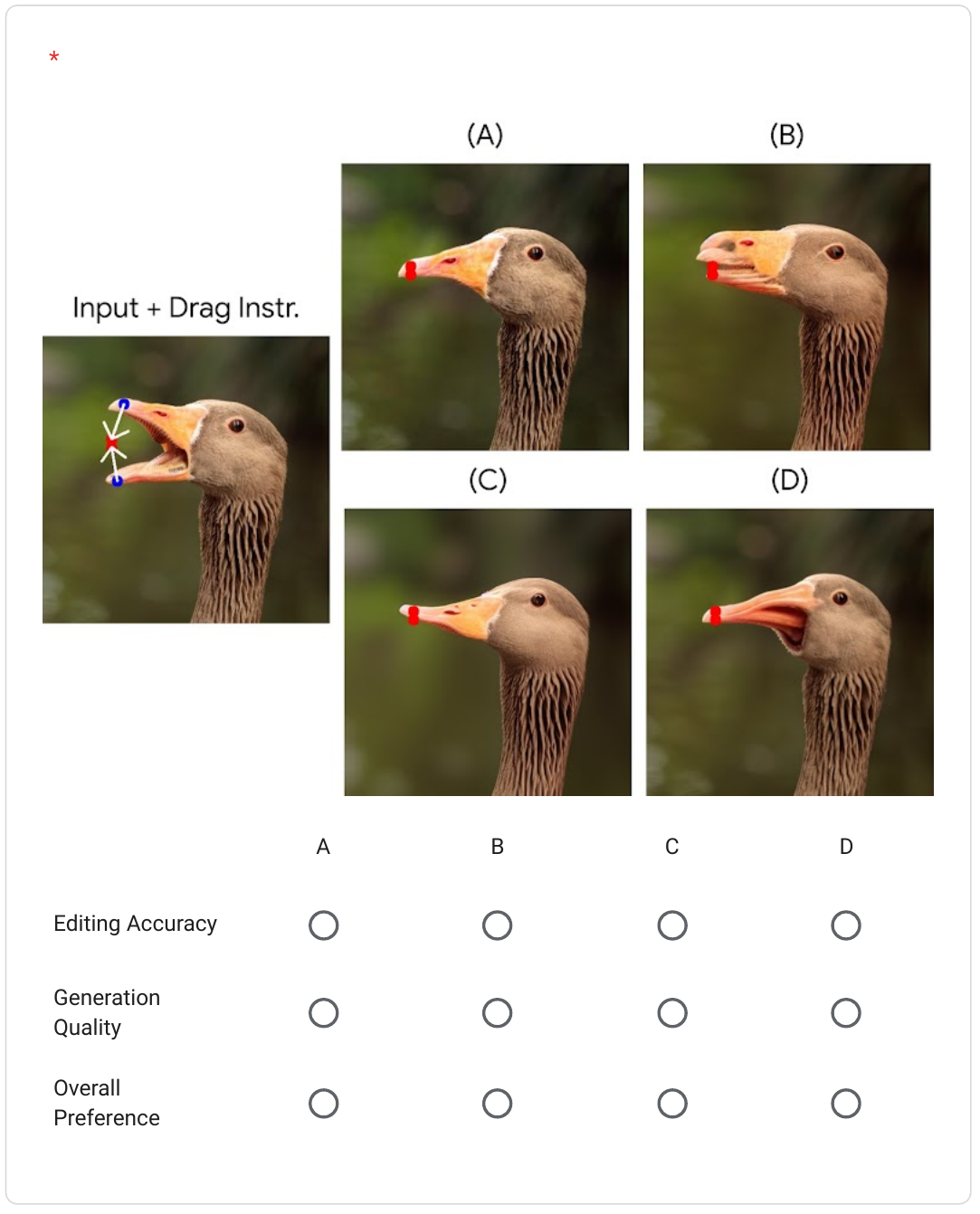}
\caption{Example question from the user study interface.}
\label{fig:user_study}
\end{figure}

Figure~\ref{fig:user_study} shows an example question from the study interface. We provide the exact instructions shown to participants below.

\begin{tcolorbox}[
    colback=gray!5,
    colframe=gray!50,
    title={User study instructions},
    fonttitle=\bfseries
]
\small
\begin{verbatim}
User Study on Drag-based Image Editing

You will be shown 20 questions with an input image, a drag instruction,
and edited results.

For each example, answer the following three questions:

Editing Accuracy:
Select the result that best follows the drag instruction. Focus on whether
image content moves from the handle point to the target point and the intended
edit is achieved.

Generation Quality:
Select the result with the best visual quality for the given edit. Focus on
image quality, realism, lack of artifacts, and preservation of regions that
should not change.

Overall Preference:
Select the result you prefer overall, considering both editing accuracy and
generation quality.

The edited results are shown in random order.
\end{verbatim}
\end{tcolorbox}

\section{Dataset Details}
\label{app:dataset}

The final dataset covers rotations (28.9\%), raising and lowering (24.3\%), translations (16.6\%), opening and closing (10.8\%), deformations (5.5\%), scaling (3.8\%), and other edits. DragBench++ image pairs are built with the same pipeline from source images that do not appear in the training data, and their drags are annotated manually. To check the edit masks provided to baselines, we measure LPIPS between each source image and its reference edit, which is 0.118 inside the mask and 0.018 outside it, so the masks cover the edited regions. Table~\ref{tab:pipeline} summarizes the model and key settings of each stage of the pipeline.

\begin{table}[!ht]
\centering
\small
\caption{Stages of the dataset construction pipeline.}
\label{tab:pipeline}
\begin{tabular}{>{\raggedright\arraybackslash}p{0.14\linewidth} >{\raggedright\arraybackslash}p{0.16\linewidth} >{\raggedright\arraybackslash}p{0.58\linewidth}}
\toprule
Stage & Model & Output and key details \\
\midrule
Source prompts & Qwen3-VL & One-sentence scene prompt of 45--110 words from a subject and a location drawn from a pool. The subject is dominant, object parts appear with their parent object, and 8--14 location details vary the scene. \\
\midrule[0.2pt]
Source images & Qwen-Image, FLUX.2, Z-Image & One image per prompt, square in about 70\% of cases and otherwise landscape or portrait, as chosen by the prompt model. \\
\midrule[0.2pt]
Edit proposal & Qwen3-VL & Five edit prompts per source image, grounded in visible parts and achievable with a few drags: articulation (\eg, opening a door, raising a limb), repositioning of an object, or deformation of a boundary. Insertion, removal, and style, lighting, or material changes are excluded. \\
\midrule[0.2pt]
Target images & FLUX.2 & One edited image per source image and edit prompt. \\
\midrule[0.2pt]
Filtering & Qwen3-VL & Binary decision on each pair. A pair is kept only if the edit is clearly achieved in the intended direction, identity and unrelated content are preserved, the change is local, and there are no artifacts, anatomy or mechanics violations, or global color and lighting shifts. \\
\midrule[0.2pt]
Keypoints & Gemini 3 Pro & One to seven keypoints on moved and anchor parts, located in both images on a 0--1000 grid, preferring corners, tips, and joints. Pairs judged unsuitable receive no keypoints and are discarded, as are drags shorter than 32 grid units. \\
\midrule[0.2pt]
Verification & DINOv3, manual & Feature consistency in local windows around each handle and target point, followed by manual inspection of the source image, target image, and drags. \\
\midrule[0.2pt]
Reasoning & Gemini 3 Flash & One paragraph of 50--200 tokens per edit direction. It describes each drag and its local effect, relates drags that act together, and ends with a sentence stating what changes and what stays the same. Coordinates are written as placeholders and filled in at training time. \\
\bottomrule
\end{tabular}
\end{table}

\section{Limitations and Future Work}

While \nickname\,achieves strong drag-based editing performance across both standard and challenging scenarios, several limitations remain. First, the reasoning mode introduces additional inference latency due to autoregressive reasoning generation, and its benefit is concentrated in ambiguous multi-drag edits. Although direct editing remains available and already performs strongly, future work could study distillation or adaptive reasoning strategies that invoke reasoning only when the edit is ambiguous. Second, our reasoning is expressed in language space, which provides interpretability but may not fully capture fine-grained visual transformations. A promising direction is to explore reasoning directly in latent or pixel space, where intermediate representations could encode spatial transformations more explicitly. Finally, although DragBench++ expands evaluation toward more semantic and ambiguous edits, drag-based editing remains under-constrained, and automatic metrics only partially reflect human preference. Future benchmarks could develop more robust evaluation protocols for settings where multiple valid edits can satisfy the same instruction.

\end{document}